\documentclass[12pt]{article}
\newcommand{\blind}{0} 
\newcommand{\titl}{Inducing Sustained Creativity and Diversity in Large Language Models}
\newcommand{\authr}{}
\usepackage[reqno]{amsmath}
\usepackage{algorithm}
\usepackage{algorithmic}
\usepackage{amssymb,amsthm,graphicx,subcaption, verbatim,url,verbatim,longtable,vmargin,placeins,
  accents,bbm,times,dcolumn,booktabs,setspace,soul,latexsym,wasysym,
  titling,enumitem,longtable,booktabs,xr,caption,mathtools, array,multirow}
\allowdisplaybreaks
\usepackage[export]{adjustbox}
\usepackage[dvipsnames,usenames]{xcolor}
\definecolor{spot}{rgb}{0.6,0,0}
\usepackage{tikz}
\usepackage{url}
\usepackage{caption}
\usepackage{soul}

\usetikzlibrary{matrix, positioning, shapes.geometric, calc}

\usepackage[all]{xy}
\setpapersize{USletter} \topmargin=0in
\newcolumntype{.}{D{.}{.}{-1}}\newcolumntype{d}[1]{D{.}{.}{#1}}
\graphicspath{{./img/}}
 
\usepackage[titletoc,title]{appendix}
\usepackage{tikz}

\theoremstyle{definition}

\theoremstyle{plain}

\if1\blind
\title{\titl}
\renewcommand{\authr}{}
\fi
\usepackage[pdftex, bookmarksopen=true, bookmarksnumbered=true,
  pdfstartview=FitH, breaklinks=true, urlbordercolor={0 1 0},
  citebordercolor={0 0 1}, colorlinks=true, citecolor=spot,
  linkcolor=spot, urlcolor=spot, pdfauthor={\authr},
  pdftitle={\titl}]{hyperref}
\usepackage{orcidlink}

  \if0\blind
\title{\titl}
\author{Queenie Luo\thanks{Ph.D. candidate, Department of East Asian Languages and Civilizations, Harvard University; sites.harvard.edu/QueenieLuo, QueenieLuo@g.harvard.edu. \orcidlink{0009-0004-1854-7968}\,0009-0004-1854-7968}\and Gary King\thanks{Corresponding Author. Albert J.\ Weatherhead III University Professor, Institute for Quantitative Social Science, Harvard University; GaryKing.org, King@Harvard.edu. \orcidlink{0000-0002-5327-7631}\,0000-0002-5327-7631}\and Michael Puett\thanks{Walter C. Klein Professor of Chinese History and Anthropology, Harvard University. \orcidlink{0009-0002-2865-0112}\,0009-0002-2865-0112}\and Michael D. Smith\thanks{John H. Finley, Jr. Professor of Engineering and Applied Sciences, Harvard University.}}

\fi

\date{}

\makeatletter
\renewcommand{\@maketitle}{%
  \newpage
  \null
  \vskip 2em%
  \begin{center}%
    \let \footnote \thanks
    {\LARGE \@title \par}%
    \vskip 1.5em%
    {\fontsize{13.3}{16.2}\selectfont
      \lineskip .5em%
      \begin{tabular}[t]{c}%
        \@author
      \end{tabular}\par}%
  \end{center}%
  \par
  \vskip 1.5em}
\makeatother

\begin{document}
\maketitle\thispagestyle{empty}\setcounter{page}{0}
\vspace{-2\baselineskip}

\begin{abstract}
  \noindent  We address a not-widely-recognized subset of exploratory search, where a user sets out on a typically long ``search quest'' for the perfect wedding dress, overlooked research topic, killer company idea, etc. The first few outputs of current large language models (LLMs) may be helpful but only as a start, since the quest requires learning the search space and evaluating many diverse and creative alternatives along the way. Although LLMs encode an impressive fraction of the world's knowledge, common decoding methods are narrowly optimized for prompts with correct answers and thus return mostly homogeneous and conventional results. Other approaches, including those designed to increase diversity across a small set of answers, start to repeat themselves long before search quest users learn enough to make final choices, or offer a uniform type of ``creativity'' to every user asking similar questions. We develop a novel, easy-to-implement decoding scheme that induces sustained creativity and diversity in LLMs, producing as many conceptually unique results as desired, even without access to the inner workings of an LLM's vector space. The algorithm unlocks an LLM's vast knowledge, both orthodox and heterodox, well beyond modal decoding paths. With this approach, search quest users can more quickly explore the search space and find satisfying answers.
 
\end{abstract}

\section{Introduction}\label{intro}

We study the \emph{search quest}, a pervasive, fundamental human activity that has not been widely recognized as a general problem to optimize for. Think of a bride-to-be determined to find the perfect wedding dress, with no more than a vague idea of what she wants at the outset. She goes online to look for inspiration, visits dress stores, talks to friends and relatives, saves images, and spends many hours exploring with search engines and LLMs. Instead of choosing from a pre-determined decision tree, or aiming to find a known target, she invents or discovers what she wants as she explores. She changes preferences as she sees dresses with features and styles she did not know existed, including color, style, price, sleeves and other parts that can be removed for dancing, scarfs, prints, veils, neck lines, silhouettes, fits, train lengths, and others separately and in interaction. Eventually she understands enough of the space of possibilities and what she likes, and she makes a choice.

A search quest for the perfect wedding dress is functionally equivalent to those we all perform regularly to find the right research topic, startup name, product design, school for your children, story idea, travel destination, career, art work for the living room wall, and many others. Search quests, more generally, have at least three characteristics: (1) the user's ultimate goal is ill defined at the outset, with only the general contours known ex ante; (2) finding the right answer is of considerable importance to users, who are willing to put in considerable time before making a decision; and (3) both the journey and the destination are essential, in that successful completion requires understanding the search space, learning or developing possibly unique preferences, and only then making a choice.

If you are searching for paper ideas, an LLM will give you a better list than you could on your own, but everyone else asking similar questions of an LLM will be given roughly the same list \cite{hao2026artificial, jiang2025artificial, Doshi2024}. This occurs because current LLMs are designed to converge to the single ``correct'' or conventional answer. Factors contributing to this pattern include common decoding methods (e.g., top-$k$ selecting high-probability tokens \cite{fan-etal-2018-hierarchical}), post-training alignment (e.g., RLHF favoring majority vote \cite{kirk2024rlhf}), and popular leaderboards (e.g., \cite{chiang2024chatbotarena}, \cite{phan2025humanitysexam}) that emphasize accuracy and majority preference. In contrast, search quest users need to learn the full space of possible answers to decide how conventional or contrarian they wish to be, and in what ways. Metrics intending to pick up ideas like ``utility,'' ``quality,'' and ``usefulness'' are subjective and so need to be evaluated by individual users, not by the majority behaviors of previous users.

In practice, tools designed for search quests need to generate creative and diverse results that are (1) \emph{relevant}, meaning within the search space defined by the prompt, (2) \emph{diverse}, meaning answers that are conceptual distinct, (3) \emph{creative}, in the specific sense that each answer (or groups of answers) covers a different region of the search space, far from others, collectively mapping out large parts of the whole space, and (4) \emph{sustained}, meaning it can provide as many nonduplicative (creative and diverse) answers as necessary for the user to learn the search space and arrive at their choice.\footnote{A mermaid-style dress and a ball gown are conceptually distinct and so are diverse, but a dress made of fiber optics is creative because it is unconventional and far from more common dresses, yet still relevant because it is in the search space of dresses. Creativity requires diversity but diversity does not require high levels of creativity.}

As we describe in Section \ref{existing}, many tools are used during search quests, but none are optimized for this purpose. Most wind up frustrating users, leading them to sift through large numbers of repetitive, homogeneous options while they try in vain to understand the broader space of possibilities. As existing search engines and LLMs get better at giving the correct answers to factual questions or reasoning tasks, their performance on search quests degrade further. Specialized algorithms designed to increase diversity and creativity for small batches of outputs accomplish the goals for which they were designed, but do not solve the search quest problem because generating larger numbers of outputs produce duplication rather than diversity and do not represent more distant, creative parts of the search space.

Since almost all people set out on search quests at some point, and usually for goals of considerable personal importance and meaning, we call on the academic community to begin to build algorithms to optimize for this goal. To help spark this research, we introduce \emph{Recoding-Decoding} (RD), a novel, easy-to-implement decoding method for accessing the rich array of creative, unconventional, contrarian, and heterodox human knowledge encoded in LLMs far from the mode, but hidden from users by standard (modal) decoding schemes (see Section \ref{rd}). RD directs generation toward less traversed but still meaningful regions of the model's knowledge space without in a way that can be easily adapted with any LLM to elicit diverse knowledge (with examples we provide for GPT-5.1 and Gemini-3). Surfacing creative, contrarian knowledge, and allowing LLMs to produce more than groupthink or repetitive answers, can make search quests more efficient and satisfying. We offer extensive empirical evaluations in Sections \ref{multip} and generalizations in Section \ref{gen}. The appendices in a separate document, along with a detailed accompanying replication dataset, provide supporting information.

\section{Existing Algorithms and Search Quests}\label{existing}

We discuss here (1) LLMs, and in particular their decoding strategies; (2) algorithms designed to improve diversity across small collections of outputs; and (3) various types of classic search and related commercial algorithms. These algorithms span fact-based and exploratory approaches \cite{marchionini2006exploratory}. Each achieves the purposes for which it was designed, and some are now employed by users on their search quests, but none satisfies their need for sustained creativity and diversity sufficient to teach users about the space of options from which they may choose.

\paragraph{Large Language Models}
LLMs encode an impressive fraction of the world's knowledge in a set of conditional probability distributions (defined over all tokens, conditioned on generated text). However, LLM developers use decoding methods to generate text optimized for the ``correct'' answer (and the fluency of generated text), meaning that they only use modal or near-modal tokens and ignore the vast majority of information encoded in their long tails (e.g., top-\emph{k} decoding selects from the \emph{k} highest-probability tokens \cite{fan-etal-2018-hierarchical}, while nucleus decoding chooses from the smallest set with cumulative probability above a threshold \cite{holtzman2020curious}). Modal decoding therefore produces homogeneous and conventional answers \cite{Xie2025, jiang2025artificial}, with upper limits on creativity \cite{cropley2025cat}, performing well below human levels in generating novel ideas and divergent thinking \cite{Ismayilzada2025, Wenger2025, wang2025large, guo-etal-2024-curious}. When writers use generative AI, individual creativity and writing quality is enhanced, but collective diversity is profoundly reduced---a potentially serious problem for universities, companies, and society at large \cite{Doshi2024}.\footnote{We inadvertently confirmed this result in a large university class when we discovered a subset of students who, despite not communicating with each other, turned in excellent essays with nearly identical arguments; upon investigation, we found they were using LLMs to help them compose essay outlines.} Although modal decoding is effective at generating correct or conventional answers, it is suboptimal for search quests.

In fact, the problem is getting worse: As LLM developers improve their models to win leaderboard competitions based on tasks with exact-match accuracy, their conditional probability distributions become increasingly peaked, causing more tail information to be ignored \cite{Divekar2024LLMvSearch}. Our experiments in Section~\ref{eval} confirm this trend that newer models generate narrower and more repetitive answers than older models. Furthermore, as the web fills up with synthetic content, web scraping turns the previously fresh LLM training into recursive training, exacerbating ``model collapse'' where even more information is relegated to the tails and thus ignored during decoding \cite{Shumailov2024, Alemohammad2024MAD, zhang2025verbalizedsamplingmitigatemode}. 

We illustrate by feeding the input sequence ``Brainstorm 5 book topics on 18th century world history.\textbackslash n1. '' to Llama2, with top predicted tokens including ``The'', ``Imp'', ``Political'', ``Age''. To understand what knowledge is encoded in these decoding paths, the complete sentence generated from each (using top-1 decoding) are European topics: ``The Age of Enlightenment'', ``Impact of the Enlightenment'', ``Political and social changes in Europe'', and ``Age of Enlightenment''. If instead we proceed further down the ranking to the $300$th--$2000$th positions, we find tokens like ``Asia'' (which extends to ``Asia's Role''), ``Second'' (which leads to ``Second Sino-Japanese War''), as well as ``African'' and ``Russian'' which point to non-European but obviously relevant world history missed by the mode. Such observations suggest a different strategy for search quests than rules designed to find a single correct answer.

\paragraph{Algorithms to Improve ``Local Diversity''}
While recent post-training and prompting strategies improve diversity for small collections of outputs (as they intend), they are not optimized for inducing the sustained creativity and diversity needed for search quests. For instance, many of these prompting methods are explicitly formulated as subset search problems or multi-stage workflows, optimized to generate a small batch of diverse outputs in a single interaction \cite{zhang2024icl, wong2025simplestrat, stay2023brainstorm, mehrotra2024associativethinking}. Though effective in enhancing diversity in a single iteration or a few outputs, ``generation quality degrades'' to less diverse or repetitive answers if extended over multiple iterations \cite{zhang2025verbalizedsamplingmitigatemode}. Post-training methods which modify loss functions to penalize homogeneity, require curating new preference datasets, meaning their creativity is bounded by the new and typically expensive training data \cite{chung2025modifying, ismayilzada2025creativepreferenceoptimization, lanchantin2025diverseoptimization}. A final approach involves selecting distant vectors in latent space representations in image models, requiring access to an LLM's internal vector space and sufficient compute resources \cite{zameshina2023diversediffusionenhancingimage}.

Existing decoding strategies for diverse text generation primarily aim to mitigate near-identical sequences generated from modal and near-modal decoding paths, and use evaluation metrics like lexical variation and sequence-level statistics, such as n-gram distinctness, repetition ratio, and entropy \cite{vijayakumar2018diverse, minh2025turning, cho2016noisyparallelapproximatedecoding, chang2025real}. In contrast, a search quest user is usually interested in conceptual diversity and unconventional knowledge (e.g. elicit creative design ideas for a storefront) where different linguistic expressions of the same concept would usually not be helpful.

\paragraph{Search-Related Algorithms}
Many other algorithms, successful for other purposes, are also suboptimal for search quests. Classic search engines, designed to satisfy users based largely on the majority behavior of previous users \cite{joachims2002optimizing}, are of limited value before search quest users have learned enough about the search space, often leading them to scour page after page of results, still unsatisfied because the list of sites quickly becomes repetitive or off-purpose. The same is true of algorithms designed for social media and other advertising-based websites attempting to keep your eyes on the page; e-commerce sites trying to get you to make purchases as fast and frequently as possible; media and streaming sites attempting to keep you engaged to reduce churn; and many others. Leading theoretical treatments, such as information foraging theory \cite{pirolli-2007-information-foraging}, are careful to discuss the existence of unsupervised learning goals but are almost entirely focused on fast and efficient fulfillment of a user's well-defined goals known ex ante.

\section{A Recoding-Decoding Algorithm}\label{rd}

For expository simplicity, we present in this section a recommended default version of our recoding-decoding (RD) algorithm, with a more general RD framework reserved for Section \ref{gen}. To access the tails of the token distribution, RD injects certain types of randomly selected tokens at selected times inside the decoding loop. This strategy diverts the model away from its modal decoding path without having to retrain or fine tune an LLM or alter its internal features.

Algorithm \ref{algor} summarizes this recommended version of RD. In each run, it introduces two forms of randomness: a random \emph{priming phrase} added to the beginning of the prompt and a random \emph{diverting token} placed at the start of each new sentence. These choices exploit LLMs' ``positional bias'' which places greater attention to tokens at the beginning and end of input sequence \cite{hsieh2024found}. RD then samples both the random priming phrase and the random diverting token and concatenates them with the generated sequence to construct the next input sentence using the LLM provider's default decoding. The diverting token randomly generated in line 5 is stored so that the same value can be used to construct the new prompt in line 6 (that uses but does not pass on its value) and the new output in line 7.
\begin{algorithm}[h]
\caption{Recoding-Decoding}\label{algor}
\begin{algorithmic}[1]
\REQUIRE User prompt $P$, token limit $N$; priming $\mathcal{V}_p$ and diverting $\mathcal{V}_d$  vocabulary sets
\STATE {\bf Function} $M(a)$: \hspace{1em}Use LLM to complete sentence continuing from text $a$
\STATE {\bf Function} $R(b)$: \hspace{1.4em}Draw element from set $b$ via uniform random sampling
\STATE {\bf Initialize:} $Y \gets \texttt{""}\qquad\qquad\qquad\qquad\mbox{ }$ (Empty string)
\WHILE{($\text{length}(Y)<N$)}
    \STATE $d \gets R(\mathcal{V}_d)$\qquad\qquad\qquad\qquad\qquad\quad(Sample and store a diverting token)
    \STATE $X \gets R(\mathcal{V}_p) + P + Y + d$\qquad\quad\quad\quad(Construct input sequence)
    \STATE $s \gets d + M(X)$\qquad\qquad\qquad\qquad\quad\,(Generate a complete sentence)
    \STATE $Y \gets Y + s$\qquad\qquad\qquad\qquad\quad\quad\,\,\,\,\,(Append sentence to growing response)
\ENDWHILE
\RETURN $Y$
\end{algorithmic}
\end{algorithm}

We construct random priming phrases by randomly selecting elements from the top 2,000 most common English nouns \cite{psobko_common_english_nouns} and, for emphasis, insert it into the phrase ``**Related to NOUN**'' (replacing ``NOUN''). We then select random diverting tokens from the three-letter starting stems of the top 5,000 common English words \cite{first20hours_google10000english}. The letter stems help divert the model onto new decoding paths restricted to a semantically appropriate subspace. For example, if we begin with the user prompt ``Brainstorm a world history book topic,'' add the random priming phrase ``**Related to FOOD:**'' at the beginning, and the random diverting token ``Pas'' to the end of the current input sequence (to begin the next sentence), the completion may become ``[Pas]ta and the silk road,'' while replacing them with ``**Related to SKY:**'' and ``Tib,'' respectively, may yield ``[Tib]etan sky burials.'' The method performs well with either component alone; examples using only the priming phrase or the diverting token are reported in Appendix~\ref{primedivert}. However, the combination of both performs best.

Implementing this algorithm requires a ``Completion API'' that allows an LLM to continue generating tokens following the input sequence. The difficulty with the more commonly available Chat Completion APIs is that they impose role labels (e.g., system, user) and a conversational context which often causes the LLM to interpret our randomly inserted tokens as typographical errors rather than continuation cues. Because Completion APIs are only available for some LLMs, we simulate it from within the standard chat interfaces or their corresponding ``Chat Completion API'' using prompt like this:
\begin{quotation}\noindent
\textbf{System prompt:} Simulate a completion API to complete the next sentence.\\
\textbf{User prompt:} \{RD modified input sequence\} (e.g. **Related to FOOD:** Brainstorm a world history topic. Pas).
\end{quotation}
Appendix~\ref{simvsreal} validates this approach using LLMs for which both types of APIs are available, and shows that both simulated and real Completions substantially increase diversity over OD, with real Completions performing the best.

We also apply a grammatical correction step (e.g., Appendix \ref{corrector}) to slightly post-process raw outputs to remove spelling errors introduced by RD interventions. Ideally, this step should serve as both a grammar and fact-corrector. However, due to LLMs' biases against unconventional content, combining it with a fact-corrector can easily revert the outputs to conventional answers, so we use it solely as a grammar corrector. Although the grammar corrector doubles token costs per run, this overhead is modest compared to reasoning models that routinely incur 10--20$\times$ tokens \cite{openai2024o1, deepseek2025r1}.

\section{Empirical Evaluations}\label{eval}

We now evaluate recoding-decoding (RD) compared to several versions of ordinary decoding (OD), ranging from the most intuitive to the most comprehensive, with respect to measures of relevance, diversity, creativity, and sustainability. Section~\ref{singlep} uses a single prompt that enables us to visualize all results in one geographic image. Section \ref{visualsummary} analyzes a small number of prompts with outputs transformed to images for easy visual comprehension. Section~\ref{multip} uses 50 highly diverse brainstorming topics across substantive fields over several LLMs, OD variants, and sustained trials and then over 500 prompts drawn from five public datasets from different substantive domains.

\subsection{Geographic Coverage of World History Battlefields}\label{singlep}

Here, we use the single prompt, ``List 5 interesting battlefields in world history,'' because the location-based responses are easy to visualize. We run this prompt 1,000 times using GPT-5.1 under both RD and OD. Both methods achieve high relevance scores (0.98 and 1.00, respectively) (see Appendix \ref{geo} for details). Figure~\ref{fig:battle} plots battlefields that appear only in RD as red dots, and those appearing in the outputs of both RD and OD as black circles. (OD identified no battlefields beyond RD.)
\begin{figure}[htbp]
    \centering
    \includegraphics[width=1\textwidth]{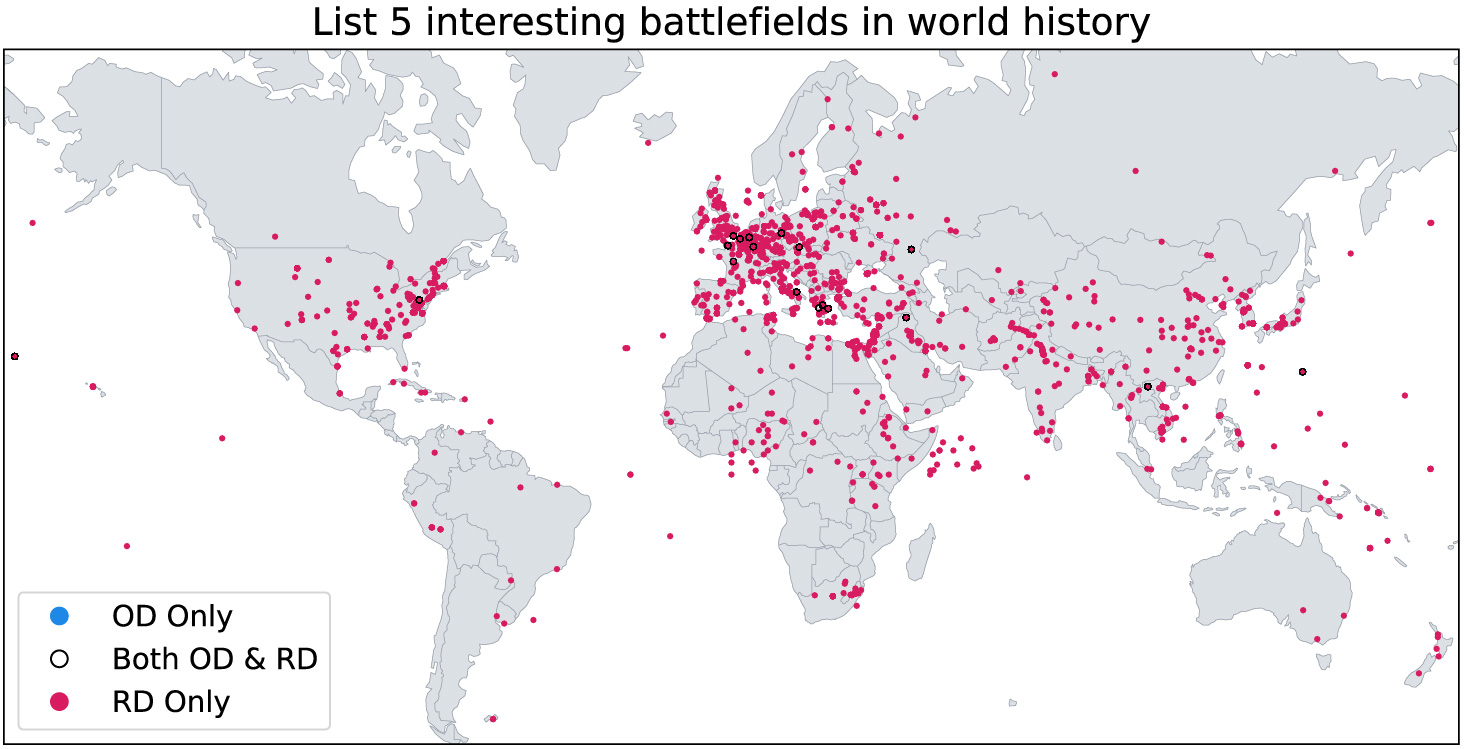}
    \caption{Geographical Distribution of Generated Battlefields. Geographic locations of battlefields generated over 1,000 runs using ``List 5 interesting battlefields in world history'' on GPT-5.1 under OD and RD. Blue dots represent battlefields appearing only in OD, red dots represent those only in RD, and black circles indicate results mentioned in both methods.}
  \label{fig:battle}
\end{figure}

The results show that OD produces only 19 unique battlefields, all in Europe and America, primarily those most famous in Western history, such as the Battles of Gettysburg, Waterloo, Stalingrad, and Marathon. In contrast, RD produces 1,307 unique battlefields, covering a much broader, and more globally distributed, geographical range, including in East Asia, South Asia, India, Russia, the Middle East, Africa, and Australia. OD is excellent at giving conventional answers, but a historian or student on a search quest for a paper topic with this prompt will find many more interesting choices using RD.

\subsection{Image-Based Evaluations}\label{visualsummary}

We now offer two intuitive evaluations based on four prompts, by turning RD and OD's textual results into images. First, we randomly draw descriptions generated by OD and RD of (a) bridal dress design ideas, (b) bouquet design ideas, and (c) Halloween party themes. We then convert these descriptions into images with Gemini-3's Nano-banana. While Nano-banana adds some randomness, the differences in diversity and creativity between RD and OD far exceed it. Both methods achieve 100\% relevance (see implementation details in Appendix~\ref{imagegeneration}; human validation in Appendix~\ref{humaneval}).

Figure~\ref{fig:gown} gives results, comparing OD (left) and RD (right) image grids across three topics. In Panel (a), OD produces largely repetitive, Western-style white bridal gowns, whereas RD yields substantially greater diversity, including personalized and culturally varied designs such as a gender-neutral jumpsuit gown, music-themed motifs, and Mongolian-inspired brocade. This aligns with the historical and contemporary role of wedding attire as an expression of cultural identity and individuality. Panels (b) and (c) show similar trends for bouquets and Halloween themes: OD generates conventional, repetitive concepts (e.g., white roses, witches, haunted mansions), while RD introduces more stylized and unconventional ideas (e.g., black roses, prismatic bouquets, bubbling cheese soup, cursed gold).
\begin{figure}[htbp]
    \centering
    \includegraphics[width=1\textwidth]{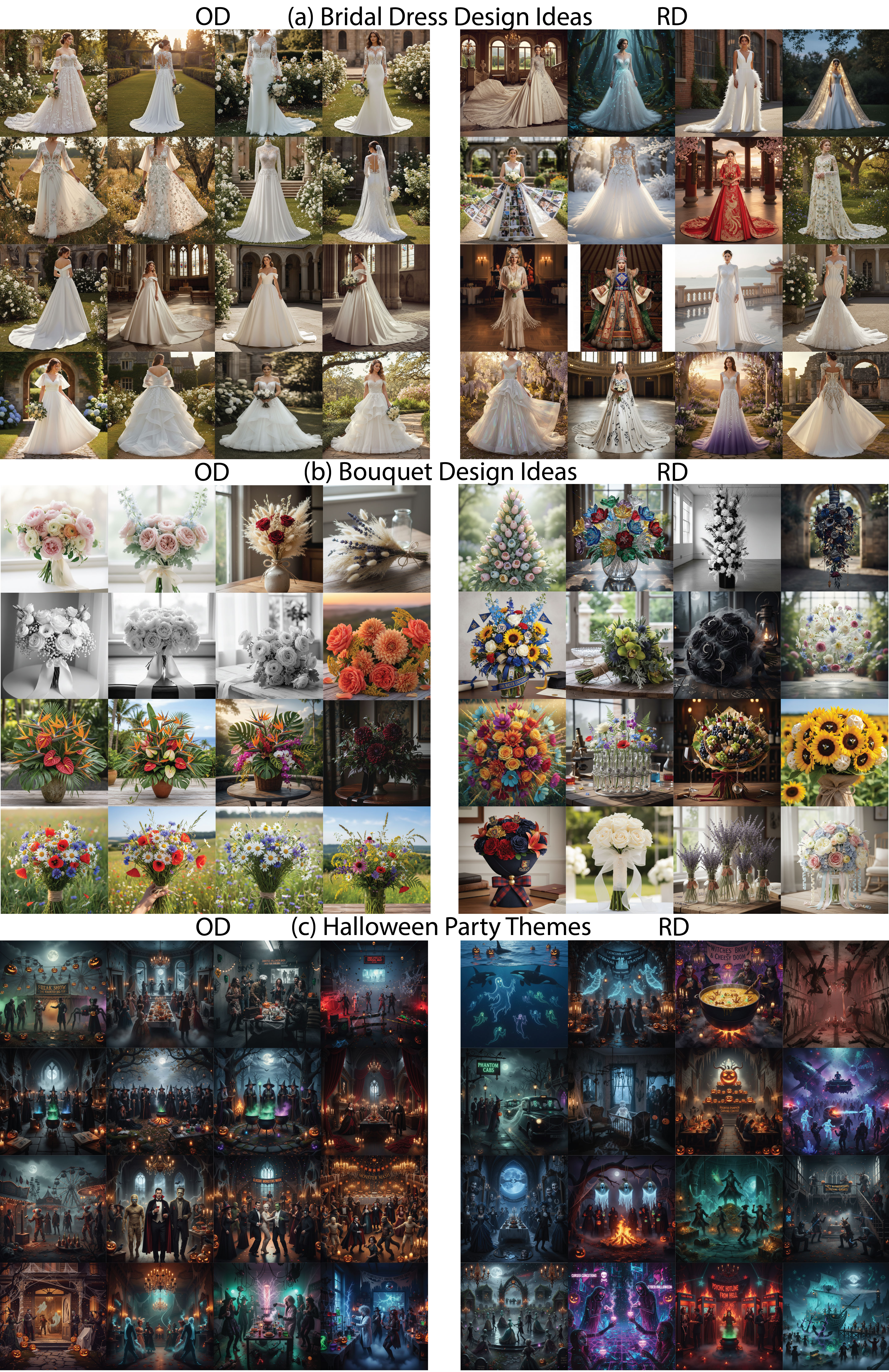}
    \caption{Multi-Prompt Visualization. Randomly sampled images each
      for OD (on the left) and RD (on the right) for three topics.
      Panel (a): Bridal dress designs, Panel (b): bouquet design
      ideas, and Panel (c): Halloween party themes.}
  \label{fig:gown}
\end{figure}

\begin{figure}[htbp]
    \centering
    \includegraphics[width=1\textwidth]{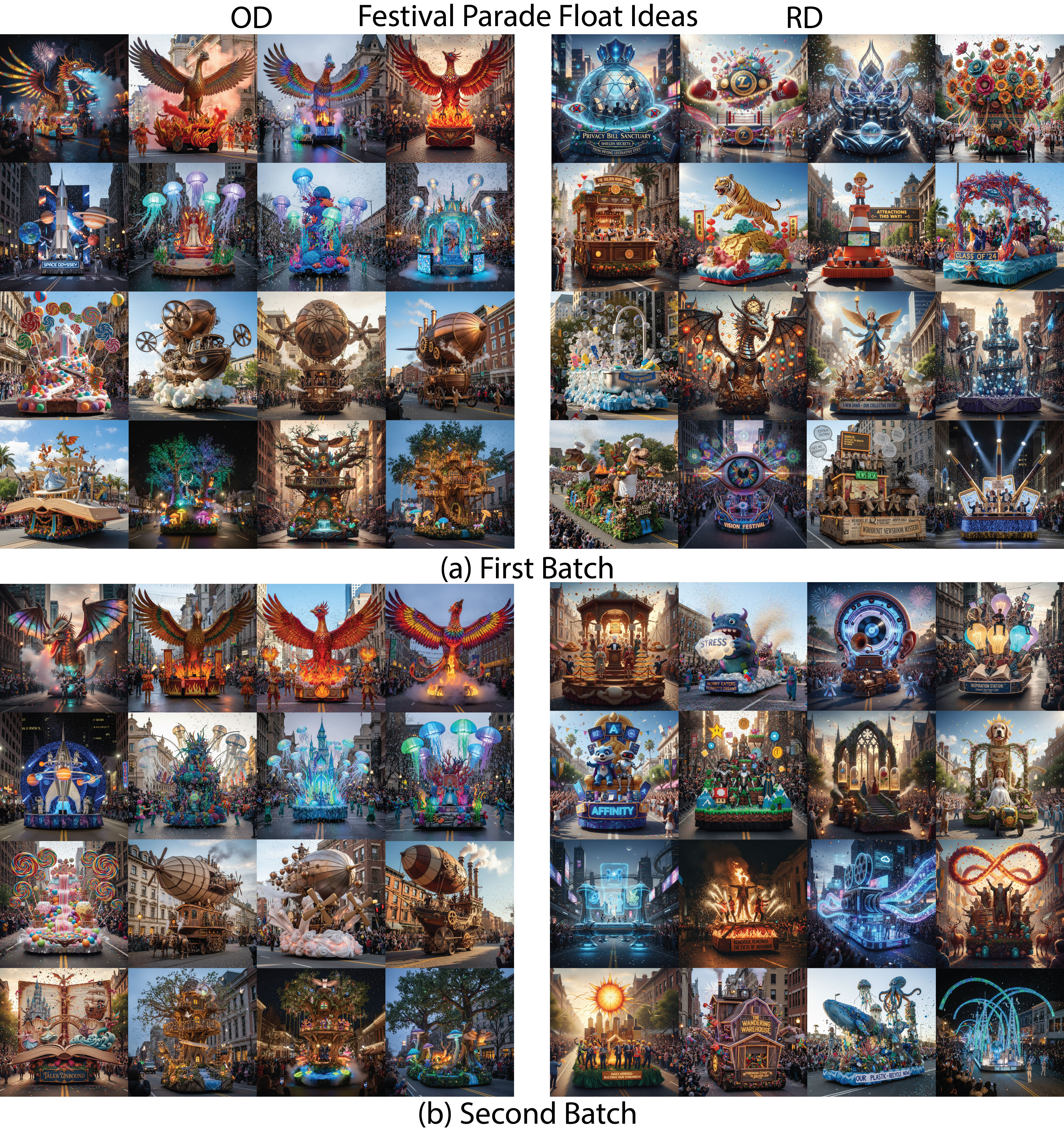}
    \caption{Visualizing Multiple Runs by Independent Users for OD
      (left) and RD (right) on GPT-5.1 in two batches in panels (a)
      and (b).}
  \label{fig:float}
\end{figure}

Finally, Figure~\ref{fig:float} extends this analysis by showing that RD also boosts collective diversity by repeating the image generation procedure twice \cite{Doshi2024}. Put differently, two \emph{independent} users are far less likely to ``show up to the same party with the same dress,'' so to speak, under RD than OD. First compare the two sets of OD in the left column of Panels (a) and (b) and note the highly similar results offered to users. For example, OD repeats nearly identical ratios of a phoenix, jellyfish, treehouse, and airships across two batches (with small image variations due to Nano-banana variance.) In contrast, RD (right) yields more varied sets to separate users: Panel (a) features traffic-cones and jurassic gardens, while (b) shows industrial ruins, Guy Fawkes bonfires, and pixelated video-game worlds. Quantitatively, RD produces 244 clusters from 250 generated ideas (50 runs × 5 ideas), whereas OD produces only 35 clusters.

\subsection{Large Scale Statistical Evaluations}\label{multip}

We now expand our scope further by comparing RD with three additional OD variants over 50 substantively diverse brainstorming topics and 500 prompts drawn from five public datasets \cite{gre,Gryphe2025WritingPrompts,zhang2025noveltybench, flux_prompts, askhistorians}. We measure ``diversity'' or conceptual distinctiveness using ten clustering algorithms, including embedding-based, graph-based, density-based, and NoveltyBench partition methods \cite{zhang2025noveltybench}. We report results using the most common embedding-based cosine similarity clustering method in the main text (see Appendix~\ref{fulltable} for results across 10 metrics). We divide creativity in two parts: We measure ``relative creativity'' by the percent of cluster centroids from one method covered by the other. If method A covers 100\% of method B but not vice versa, method A covers a broader search space than method B. We define ``absolute creativity'' as the euclidean distance of a cluster to the closest previously generated cluster centroid. This metric quantifies whether a method continues to produce clusters that are farther from, not merely distinct from, existing ones.

We compare methods based on four LLMs, listed in increasing order of performance on fact-based benchmark scores --- (1) Deepseek-3, (2) GPT-3.5, (3) GPT-5.1, and (4) Gemini-3 --- labeling ordinary decoding methods as OD1--OD4 and recoding-decoding methods as RD1--RD4. We also include four other baselines including appending (1) chat history ($\text{OD}_h$), (2) a single prompt engineering phrase ($\text{OD}_s$), (3) multiple prompt engineering phrases ($\text{OD}_m$), and (4) using temperature 1.6 with grammatical post-processing ($\text{OD}_{16}$) (the best-performing temperature in our ablation study). See Appendix~\ref{fulltable} and \ref{temperature}. Zhang et al. \cite{zhang2025noveltybench} find that keeping previous chat history and explicitly requesting different answers (equivalent to $\text{OD}_h$ in our paper)—is the most effective method among their baselines. Nevertheless, it remains less effective than RD.

\subsubsection{50 Brainstorming Topics}\label{contestresults}

Across 50 substantively different brainstorming topics, LLM-based evaluations show consistently high relevance for all methods. RD obtains relevance scores of 0.99 on GPT-3.5, GPT-5.1, and Gemini-3, and 0.94 on DeepSeek-3, comparable to OD (0.99–1.00 on four models). RD does not degrade relevance in open-ended brainstorming tasks, particularly on newer models (Appendix~\ref{tab:multiprelevance_scores}).

\begin{figure}[htbp]
    \centering
    \makebox[\textwidth][c]{\includegraphics[width=1.3\textwidth]{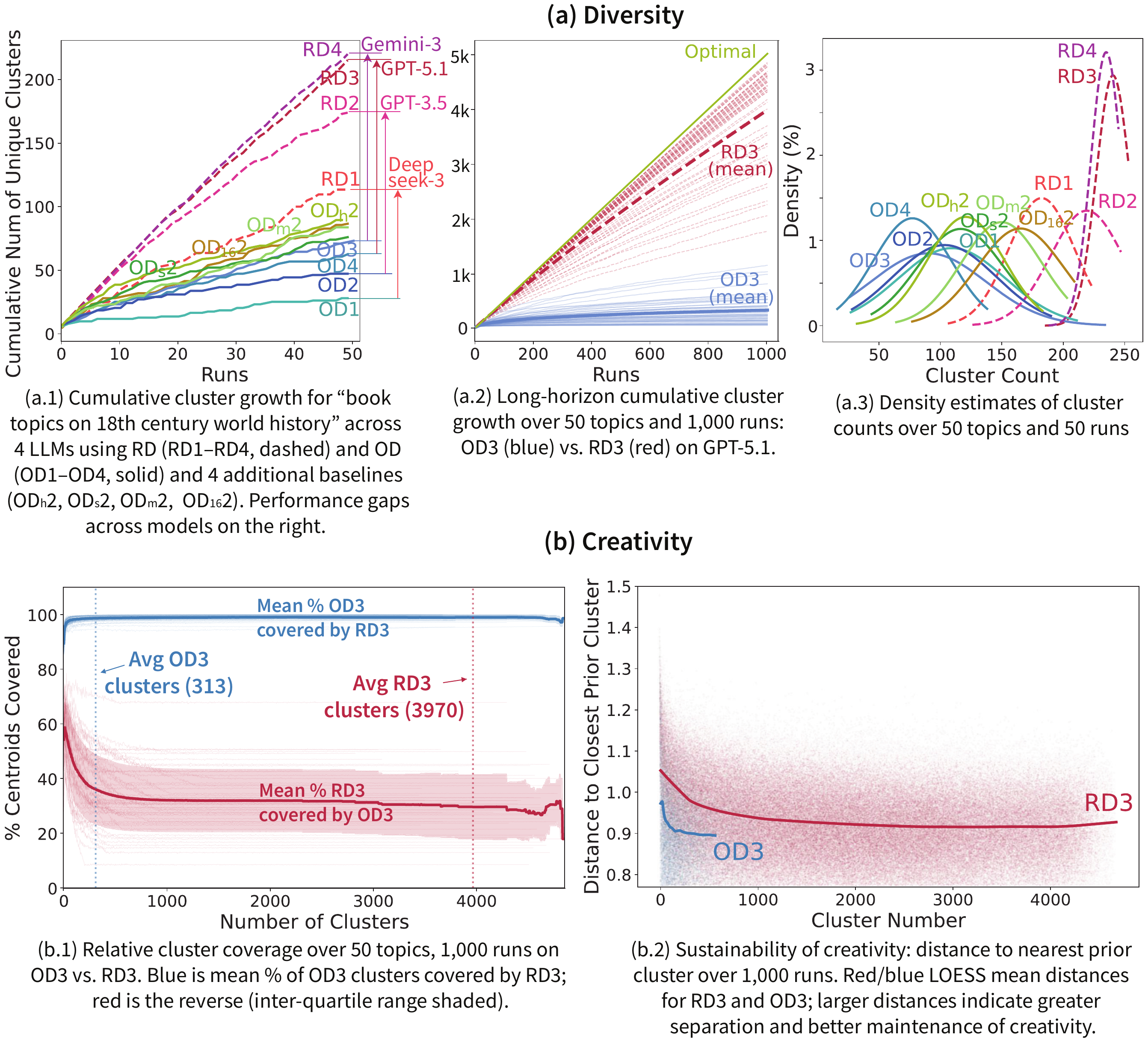}}
    \caption{Diversity and Creativity: 6 Methods, 4 LLMs. (a.1)-(a.3): \emph{Diversity}; (b.1)-(b.2): \emph{Creativity}.}
  \label{fig:contest}
\end{figure}

Figure~\ref{fig:contest} gives our (a) diversity and (b) creativity results. Panel (a.1) compares methods via cumulative cluster growth curves for one of the 50 topics (``Brainstorm 5 book topics on 18th century world history.''). On the horizontal axis is the run number, with the vertical axis representing the total number of clusters (i.e., unique ideas for world history topics). The four RD algorithms (corresponding to LLMs) appear as dashed lines and all the OD algorithms as solid lines; colors distinguish among individual algorithms.

Three results are particularly noteworthy in Figure \ref{fig:contest} (a.1). First, the dashed lines for all four RD algorithms are higher than, and thus outperform, all OD methods. Second, RD4 using the best performing and newest LLM has nearly perfect performance, where nearly every output is a unique world history topic (see the top dashed line, which is nearly linear, producing five clusters for each run of five). Third, RD based on newer, higher accuracy LLMs outperform OD algorithms based on the same LLM even more than for older LLMs. Vertical arrows at the right of panel (a.1) annotate the performance gaps between RDs and their corresponding ODs for specific models. Put differently, as LLMs perform higher on accuracy-based benchmarks for which they were designed, they perform worse at exploring the search space of diverse possible answers to open ended, nonfactual questions. Better LLMs, trained on more data, have more information encoded in their vector spaces but traditional decoding ignores more of it, as their likelihood functions are more peaked around the mode.

Panel (a.2) extends this analysis to 1,000 runs across 50 topics to evaluate sustained diversity. RD3 (red dashed line) consistently outperforms OD3 (blue solid line) across all topics and, for some, RD3's growth remains nearly linear even at the 1,000th run. Topics at the lower end of RD3, which exhibit slower growth, tend to have a finite search space, for example, ``fashion accessories,'' which has a limited set of valid answers. In contrast, the topics remain which linear at the 1,000th run, such as ``advertising campaign,'' have much larger answer spaces. Panel (a.3) summarizes the diversity analyses by plotting histograms (via density estimation) of total cluster counts across 50 topics, each over 50 runs. OD distributions concentrate at lower cluster counts (blue/green, to the left of the panel), whereas RD distributions are shifted to the right and become increasingly separated as model capability improves (pink/purple dashed curves). RD with more capable models are also more peaked, indicating higher performance across topics.

We evaluate versions of creativity in Figure~\ref{fig:contest} (b). Panel (b.1) evaluates relative cluster coverage between RD3 and OD3 over 50 topics and 1,000 runs. The blue bars represent the mean percentage of OD3 clusters covered by RD3, while the red bars indicate the reverse (with interquartile ranges shaded). The results show that RD3 covers nearly all clusters previously produced by OD3 (mean close to 100\%), while OD3 only covers about 30-40\% of RD3's clusters, consistent with RD3 exploring a much broader part of the search space. Panel (b.2) assesses sustained creativity. RD3 consistently maintains a higher distance to the nearest prior cluster centroid compared to OD3, indicating that RD3 continues to produce novel ideas over time without converging to existing clusters.

\subsubsection{500 Topics from Public Datasets}\label{largescale}

We also conduct a large-scale evaluation using five datasets (sampling 100 prompts from each): (1) NoveltyBench \cite{zhang2025noveltybench}, (2) GRE analytical writing topics \cite{gre}, (3) creative writing prompts \cite{Gryphe2025WritingPrompts}, (4) image prompt expansion \cite{flux_prompts}, and (5) r/AskHistorians \cite{askhistorians}. Results are consistent with our brainstorming evaluation (Appendix~\ref{largescale}), demonstrating that RD substantially increases diversity and creativity across a wide range of domains while maintaining comparable relevance.\footnote{Although we design RD for search quests, it performs well on ``local diversity'' too using NovelityBench's 100 prompts and independent human responses. We run each prompt 5 times under RD, OD, and Verbalized Sampling \cite{zhang2025verbalizedsamplingmitigatemode}. For diversity, we count prompts where all 5 results fall into distinct clusters, determined by embedding-based cosine similarity. On this metric, RD is 95.0\%, VS is 85.0\%, human responses are 82.0\%, and OD3 is only 15.0\%.}

\section{Generalizations}\label{gen}
\begin{figure}[htbp]
  \centering
  \includegraphics[width=0.35\textwidth]{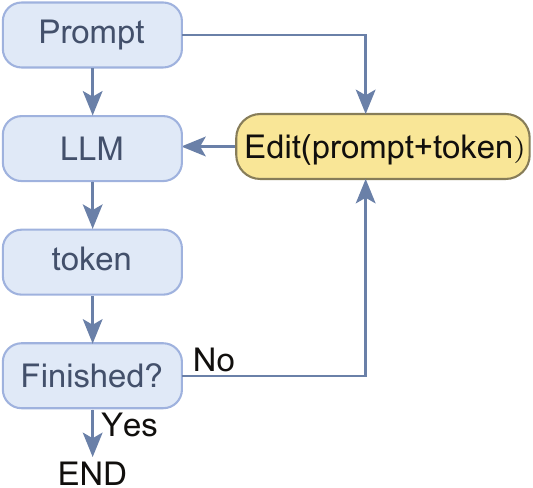}
  \caption{RD architecture. RD integrates a \emph{token-level} editor that watches the LLM output while it generates the next token. At each step, the editor decides \emph{if} something should be changed and \emph{what} to change. If needed, it can delete, replace, or add tokens, and then sends the updated text back to the LLM. If no changes are needed, it just leave the input tokens unchanged and lets the LLM continue as is.}\label{fig:RD}
\end{figure}

We now describe the general version of our RD algorithm, with RD architecture illustrated in Figure~\ref{fig:RD}. This algorithm integrates a \emph{token-level} editor watching LLM output during generation. The editor decides if and how tokens should be changed by deleting, replacing, or adding them at any point, and then sending the updated text back to the LLM to generate the next token.

Editing can be triggered at different text locations (e.g., paragraph, section boundaries, verbs, or adjectives), or handled by a neural network. Examples of what to change include injecting customized domain-specific or language-specific tokens, removing harmful speech, reducing political or ideological imbalance. Appendix~\ref{addi_exp} presents configurations covering domain-specific elicitation, opinion moderation, cultural elicitation using multilingual letter stems, and advertisement insertion.

\section{Concluding Remarks}\label{concl}
The AI community has worked intensely to increase LLM accuracy, with unprecedented funding and effort and spectacular results. Yet, this very success degrades performance on competing goals. This is especially true for the search quest, the unsupervised journey most of us regularly take to learn about and develop some of our most personally meaningful goals and decisions. Through extensive empirical evaluations, we show that our recoding-decoding algorithm substantially improves upon ordinary decoding strategies.

For future research, researchers may wish to consider (1) formalizing the search quest as a novel objective function where, in the absence of noise, the user's ultimate choice can only be determined by knowing their view  (i.e., ``potential outcomes'') of all possible results in the search space, since viewing any one may change their trajectory and ultimate choice; (2) Improving metrics; (3) Investigating novel training and inference architectures to achieve similar goals; (4) Designing RD architectures for generating high-quality synthetic data; (5) Collecting observational and experimental data to improve algorithms; and (6) developing UIs with adjustable parameters (e.g., controllable novelty levels) that support iterative efforts to help humans in their search quests \cite{marks1997design}.

\singlespace
\bibliographystyle{unsrt}
\bibliography{references.bib}

\section*{Acknowledgements}
Our thanks to Peter Bol, António Câmara, Yung-Sung Chuang, Kosuke Imai, Connor Jerzak, Mitsuru Mukaigawara, Hanspeter Pfister, Rahul Razdan, Till Saenger, and Brandon Stewart for helpful suggestions.




\clearpage
\setcounter{table}{0}
\renewcommand{\thetable}{A\arabic{table}}
\renewcommand{\thefigure}{A\arabic{figure}}
\renewcommand{\thesection}{A\arabic{section}}

\begin{appendices}
\section{Battlefield Implementation Details}\label{geo}
Here, we provide details for the analysis in Section \ref{singlep}. First, we set a maximum output token length of 150 for each run and prepend the substantive prompt ``Respond in bullet points. Do NOT include sub-bullet points.'' We then use GPT-4o with structured output to extract the battlefields mentioned. Next, we use Gemini-2.5 Flash with search retrieval to fact-check the extracted battlefields, obtain their corresponding geolocations, and remove duplicates. Overall, RD and OD produce 113 and 0 irrelevant battlefields out of 5,000 extracted ones, resulting in relevance scores of 0.98 and 1.00, respectively. We remove irrelevant battlefields before visualization.

\section{Image Generation Implementation}\label{imagegeneration}
To generate the images used in Section~\ref{visualsummary}, we first generate design ideas using four prompts running each prompt 50 times with GPT-5.1. An example full prompt is: ``Respond in bullet points. Do NOT include sub-bullets. Limit each point to 10 words. Brainstorm 5 bridal dress design ideas.'' Other prompts vary only in the main subject (e.g., replacing “bridal dress” with “bouquet”). The additional instructions ensure consistent outputs for comparison. Next, we use GPT-4o with structured output to extract the generated ideas. Because the raw outputs are already in bullet-point form, the extracted ideas are highly accurate and consistent. We also measure idea relevance using GPT-4o, under which both RD and OD achieve 100\% relevance scores on these topics. The relevance-judge prompt is provided in Appendix~\ref{llmjudge}.

To generate the images, we first cluster the generated ideas for each method using embedding-based cosine similarity. We then sample 16 ideas per method (RD and OD) in proportion to cluster sizes, ensuring the samples reflect each method's overall idea distribution. Sampling proportional to cluster sizes ensures our visualizations accurately represent each method's true output distribution \cite{cochran1977sampling}. After sampling the 16 idea descriptions, we convert each into an image prompt by attaching a fixed prefix and quality specification before sending it to Gemini-3 Nano Banana to standardize style and image quality: ``Generate an image of a \{object\}:\{sampled idea description\}. Photorealistic style. High resolution. HD''. When style and quality instructions are not provided, Nano Banana may produce images with varying styles (e.g., painting or cartoon), which makes conceptual comparisons difficult.

\section{Large Scale Statistical Evaluation Details}\label{fulltable}

We now summarize the evaluation metrics and procedures used in Sections~\ref{contestresults} and~\ref{largescale}, and report the numerical results.

\subsection{Evaluation Metrics for Relevance, Diversity, and Creativity}
We assess \emph{relevance} using an LLM-judge (GPT-4o) ~\cite{zheng2023judgingllmasajudgemtbenchchatbot} (prompt details in Appendix~\ref{llmjudge}) on a three-level scale: irrelevant (0 points), partially relevant (1 point), and relevant (1 point). We sample 20 outputs for each model-prompt-method setting. Although we distinguish between ``partially relevant'' and ``relevant'' during LLM evaluation, both categories receive 1 point. This design addresses a well-documented bias in LLM judges and LLM-based evaluators that systematically favors typical, mainstream responses over equally valid but unfamiliar alternatives due to typicality bias in their training data~\cite{kirk2024prism, sorensen2024roadmap}. For example, when evaluating responses to ``Tell me a world history book topic,'' GPT-4o rates ``medieval Church funerary practices'' as ``Relevant'' but ``Tibetan sky burials'' as ``Partially Relevant'' despite both being important religious funerary practices.

We further verified the alignment between the LLM judge and human judges on a sample of 500 responses and found 80\% majority-vote agreement on a three-level scale (Relevant, Partially Relevant, Irrelevant), and 97\% agreement after merging ``Relevant'' and ``Partially Relevant'' into a single category (see Appendix~\ref{humaneval} for details). Our human evaluation (Appendix~\ref{humaneval}) demonstrates that of the 108 responses that the LLM judge rated as merely ``Partially Relevant,'' a human majority vote classified 77\% of them as fully ``Relevant.'' This discrepancy further underscores the LLM judge's bias against unconventional or less-dominant ideas.

We quantify \emph{diversity} by first extracting ideas using GPT-4o with structured output (which is highly accurate because outputs are arranged in bullet points). Next, we compute the cumulative number of unique clusters generated at each run. That is, by running the clustering algorithm on all responses produced up to a given point (Run \emph{i}), we record the unique cluster count and plot this growth against the run index. We apply 10 distinct clustering algorithms. As we observe high consistency across them all, we present in the text only the most commonly used metric, embedding-based cosine similarity. We also employ an LLM-judge to evaluate pairwise diversity. We sample 20 response pairs within each model-prompt-method combination and score them on a three-point scale: almost identical (0), partially similar (1), and mostly different (2). The mean score across 50 topics serves as the method's diversity score.

Our clustering methods include (1) Embedding-based Cosine Similarity; (2) TF-IDF (Term-Frequency-Inverse-Document-Frequency)-based Cosine Similarity; (3) Embedding-based Hierarchical Agglomerative Clustering (HAC) A; (4) Embedding-based Hierarchical Clustering B; (5) Graph-based Spectral Clustering; (6) Leiden Community Detection; (7) BIRCH (Balanced Iterative Reducing and Clustering using Hierarchies); (8) DeBERTa model for similarity classification from NoveltyBench \cite{zhang2025noveltybench}; (9) BERTscore partition from NoveltyBench \cite{zhang2025noveltybench}; (10) Unigram partition from NoveltyBench \cite{zhang2025noveltybench}. We apply the same clustering threshold 
$\tau$ across all methods to ensure a fair comparison. The value of $\tau$ is chosen carefully so that it meaningfully differentiates performance between methods. Code is in our replication archive.

We define \emph{creativity} as the degree to which one distribution of ideas occupies novel regions in the embedding space not covered by the other. To measure this, we first cluster each set of response embeddings using cosine-similarity clustering with a threshold of $0.73$ for brainstorming tasks and $0.83$ for the five public datasets: embeddings that are sufficiently similar are grouped together, and each cluster is represented by its centroid. We then assess \emph{coverage} between two sets of centroids using a nearest-neighbor criterion. A centroid from one distribution is considered ``covered'' by the other if its nearest neighbor in the other set falls within an adaptive distance threshold, set as the 95th percentile of the internal nearest-neighbor distances within the reference set. This adaptive threshold naturally accounts for differences in cluster density across prompts and models, avoiding the need to hand-tune a fixed distance cutoff. We report the percentage of covered centroids, averaged over 50 bootstrap iterations at each evaluation point. The shaded region in the plot represents the interquartile range (25th--75th percentile) across prompts, indicating where the middle 50\% of individual coverage curves fall.

We do not use metrics like ``quality'' or ``utility'' because these metrics are designed for different purposes and happen to be misaligned with search quest goals. For example, NoveltyBench's quality metric relies on the Skywork-Reward model \cite{liu2024skywork}, which is intended to pick up mainstream human preferences, and so is useful for different purposes. Cognitive psychology suggests that humans systematically favor familiar more easily-processed content through mechanisms such as mere-exposure effects~\cite{zajonc1968attitudinal}, availability heuristics~\cite{tversky1973availability}, and processing fluency~\cite{reber2004processing}. For the prompt ``Tell me a world history book topic,'' as the Skywork-Reward model is aligned with human preferences, it assigns 2/10 to ``Tibetan sky burial in Tibet'' but 10/10 to ``The Enlightenment in Europe,'' despite sky burials being a significant religious funerary practice in Tibet. Because such topics are less prominent in Western discourse, mainstream-aligned models systematically assign them lower quality scores. A similar pattern appears for the prompt ``Brainstorm a wedding dress design idea,'' where ``Modern romantic french lace gown with a train'' receives 6/10 quality score, but ``Gender-neutral jumpsuit with a train for a lesbian bride'' receives only 1/10. Quality metrics such as these systematically undervalue contrarian or unconventional responses~\cite{meyers1989schema}. Mainstream ideas are useful to be aware of, but for search-quest users seeking varied perspectives or overlooked ideas, repeating mainstream ideas can be counterproductive. Quality metrics aligned with mainstream views are not suitable for evaluating search quest responses.  Utility evaluation defined by conventional standards of ``usefulness'' is also not suitable for judging search quest responses, as the notion of ``utility'' itself is highly subjective and context-dependent. Similarly, LLM-based judges, trained of course on human discourse, tend to favor mainstream answers when comparing responses~\cite{zhang2025verbalizedsamplingmitigatemode}.

\subsection{Procedures used in Section~\ref{contestresults}}
We expand the 4 brainstorming topics into 50 highly distinct ones spanning a wide range of domains, including history book topics, museum exhibition, and video game concepts. The prompts are also prepended with formatting instruction described in Appendix~\ref{imagegeneration}.

We perform 50 runs for each of 50 topics across 6 methods, with a maximum of 150 tokens per run (enough for a list of five ideas). We use a temperature of 1.4 for Deepseek-3 \footnote{At $T=1.0$, Deepseek-3 produces highly repetitive outputs (99\% token overlap across runs), so we use $T=1.4$ for both RD and OD. LLMs exhibit different probability distributions at the same temperature \cite{meister2023natural}.} and 1.0 for the other three models, leaving all other parameters at their defaults.

Our evaluation is run on these methods: \emph{Recoding-Decoding (RD)}: The original prompt repeated for 50 or 1,000 runs using the configuration described in Section~\ref{rd}; \emph{Ordinary Decoding (OD)}: The original prompt repeated for 50 or 1000 runs; \emph{Ordinary Decoding with chat history ($\text{OD}_h$)}: The original prompt used in the first run, followed by 49 runs that include all prior chat history with the additional prompt to ``Generate 5 more ideas''; \emph{Ordinary Decoding with a single engineered phrase ($\text{OD}_s$)}: The phrase ``Think outside the box. '' appended to the end of the original prompt and used for 50 runs; \emph{Ordinary Decoding with multiple engineered phrases ($\text{OD}_m$)}: We created 50 prompt-engineering phrases (e.g., ``Think outside the box,'' ``Please be diverse,'' and ``Give me unknown ideas''; details  appear in our replication archive) and prepend one of these of the original prompt.

\subsection{Procedures used in Section~\ref{largescale}}
We randomly sample 100 prompts from (1) NoveltyBench \cite{zhang2025noveltybench}, (2) GRE analytical writing topics \cite{gre}, (3) creative writing prompts \cite{Gryphe2025WritingPrompts}, (4) image prompt expansion \cite{flux_prompts}, and select 100 open-ended questions from r/AskHistorians \cite{askhistorians}, excluding narrowly focused and fact-based questions (e.g., ``When was the Eiffel Tower built?'').

For our evaluation of 500 topics from five datasets in Section \ref{largescale}, we evaluate RD using GPT-3.5, GPT-5.1, and Gemini-3 (pro). We select GPT-3.5 because this model is accessible through their Completion APIs. In addition, we incorporate GPT-5.1 and Gemini-3 to demonstrate RD's effectiveness on newer models by using the simulated Completion API. For each model-prompt-method setting, we perform 50 runs with a maximum of 300 tokens per run.

\begin{figure}[htbp]
    \centering
    \includegraphics[width=\textwidth]{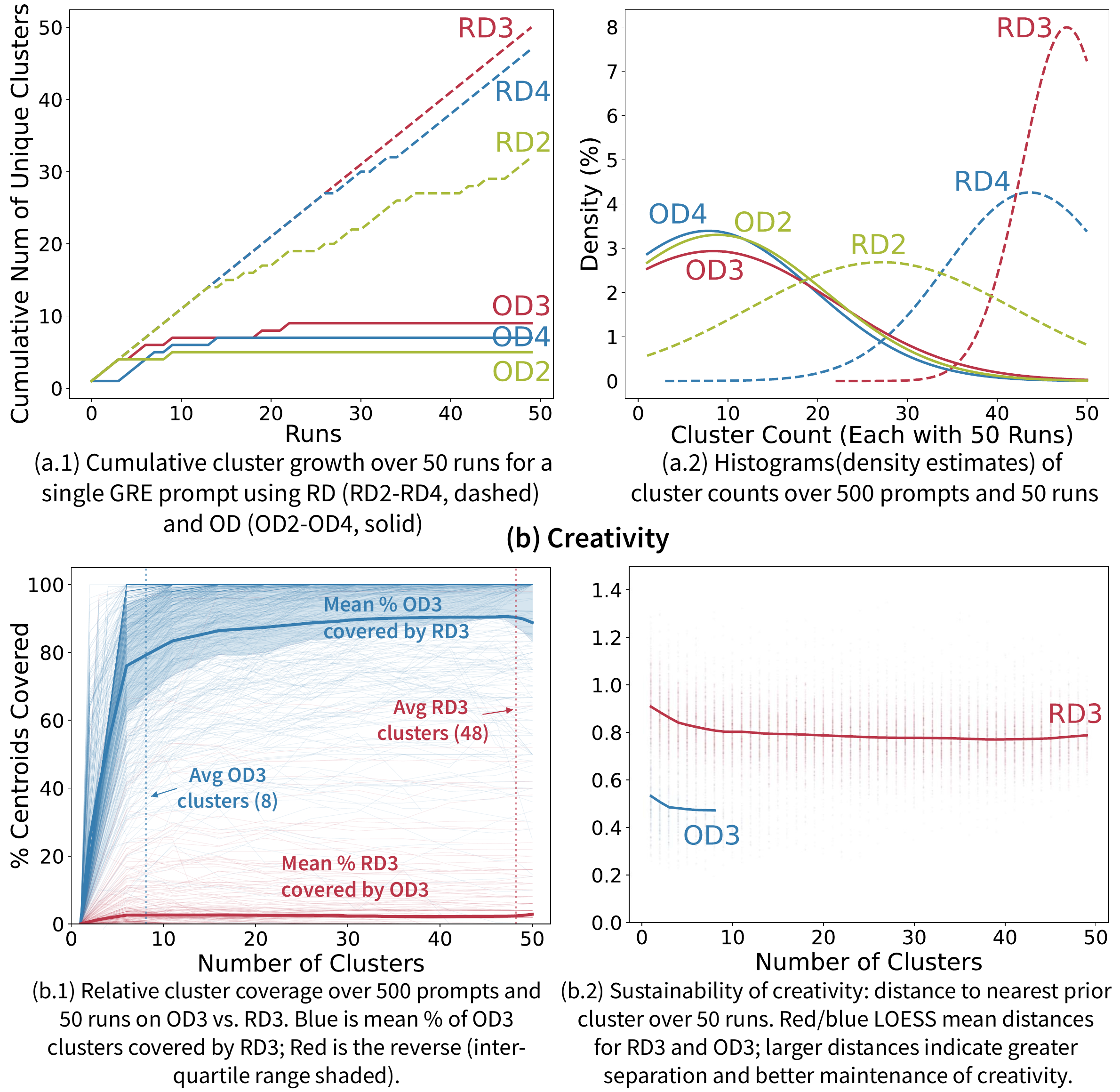}
    \caption{Diversity and Creativity: 5 Datasets (500 Prompts), 3 LLMs. (a.1)-(a.2): \emph{Diversity}; (b.1)-(b.2): \emph{Creativity}.}
    \label{fig:utility_dissimilarity}
\end{figure}

Figure~\ref{fig:utility_dissimilarity} compares RD and OD on diversity and creativity, as above. Panel (a.1) shows cumulative cluster growth for a single GRE prompt ``Scientists and other researchers should focus their research on areas that are likely to benefit the greatest number of people.'' Panel (a.2) presents histograms (via density estimation) of total cluster counts over all 500 prompts (100 from each of the four datasets) and 50 runs. Similar to the results in Section~\ref{contestresults}, all dashed lines (RD2-RD4) outperforms solid lines (OD2-OD4); RD3 (GPT-5.1) has perfect performance over 50 runs; RD's advantage over OD increases on newer models.

Panel (b.1) evaluates relative cluster coverage between RD3 and OD3 over 50 topics and 1,000 runs. The blue bars represent the mean percentage of OD3 clusters covered by RD3, while the red bars indicate the reverse. The interquartile ranges are shown as shaded areas. RD3 covers 80-90\% of the clusters found in OD3 as the number of clusters increases, suggesting that RD3 captures nearly all the semantic space explored by the baseline. In contrast, OD3 only covers a negligible fraction (near 0\%) of the RD3 clusters. This massive disparity indicates that RD3 explores a significantly larger conceptual space. Panel (b.2) measures sustained creativity by calculating the euclidean distance to the closest prior cluster centroid as more outputs are generated. RD3 consistently maintains a higher distance to the nearest prior cluster (averaging around 0.8) compared to OD3 (averaging around 0.5) and stop continuing generating more cluster around 8 clusters. While both models show a slight initial dip, RD3 continues to produce novel clusters over long-horizon.

\subsection{Numerical results}

\begin{table}[htbp]
\centering
\caption{LLM-Rated Diversity and Relevance Scores across 10 Methods over 50 Brainstorming Topics}
\label{tab:scores}
\begin{tabular}{lcc}
\toprule
\textbf{Method} & \textbf{Diversity} & \textbf{Relevance} \\
\midrule
OD1  & 0.6880 & 0.9930 \\
OD2  & 0.6410 & 0.9970 \\
OD3  & 0.5910 & 0.9830 \\
OD4  & 0.4688 & 0.9960 \\
$\text{OD}_h2$ & 0.9077 & 0.9810 \\
$\text{OD}_m2$ & 0.8115 & 0.8650 \\
$\text{OD}_s2$ & 0.6730 & 0.9950 \\
$\text{OD}_{16}2$ & 0.8054 & 0.9200 \\
RD1  & 0.9520 & 0.9454 \\
RD2  & 0.9425 & 0.9965 \\
RD3  & 0.9795 & 0.9911 \\
RD4  & 0.9593 & 0.9790 \\
\bottomrule
\end{tabular}
\label{tab:multiprelevance_scores}
\end{table}

Table~\ref{tab:multiprelevance_scores} shows LLM-rated diversity and relevance across 50 brainstorming topics. RD variants consistently achieve substantially higher diversity scores (ranging from 0.94 to 0.98) compared to standard OD methods (0.47 to 0.69). Notably, RD2, RD3, and RD4 achieve diversity scores above 0.94 while preserving high relevance scores. Among the OD variants, $\text{OD}_m2$ achieves the highest diversity score at 0.81, but this comes at a significant cost to relevance (0.863). In contrast, other OD variants such as $\text{OD}_h2$ maintain relevance above 0.98 but fail to match the diversity gains of RD methods. These results indicate that while certain OD modifications can improve diversity, they either sacrifice relevance ($\text{OD}_m2$) or fall short of the diversity levels achieved by RD, which consistently balances both metrics effectively.

\begin{table}[!ht]
\centering
\setlength{\tabcolsep}{1mm}{
\begin{tabular}{clcc}
\toprule
\textbf{Dataset} & \textbf{Method} & \textbf{Diversity} & \textbf{Relevance}\\
\midrule
\multirow{2}*{r/AskHistorian} & RD & \textbf{65.18} & 98.48\\
 & OD & 42.24 & 99.91\\
\midrule
\multirow{2}*{NBwildchat} & RD & \textbf{63.03} & 97.15\\
 & OD & 46.12 & 98.35\\
\midrule
\multirow{2}*{GRE} & RD & \textbf{50.45} & 98.23\\
 & OD & 31.35 & 100.0\\
\midrule
\multirow{2}*{Image Prompt Expansion} & RD & \textbf{79.86} & 99.38\\
 & OD & 57.58 & 99.64\\
\midrule
\multirow{2}*{r/Writing} & RD & \textbf{81.57} & 99.31\\
 & OD & 53.05 & 99.85\\
\bottomrule
\end{tabular}}
\caption{Performance evaluation of the GPT-3.5 model. We compare RD against OD across five datasets (500 prompts).\label{tab:gpt}}
\end{table}

\begin{table}[!ht]
\centering
\setlength{\tabcolsep}{1mm}{
\begin{tabular}{clcc} 
\toprule
\textbf{Dataset} & \textbf{Method} & \textbf{Diversity} & \textbf{Relevance}\\
\midrule 
\multirow{2}*{r/AskHistorian} & RD & \textbf{63.66} & 94.84\\
 & OD & 17.62 & 99.86\\
\midrule 
\multirow{2}*{NBwildchat} & RD & \textbf{73.18} & 96.15\\
 & OD & 31.0 & 99.6\\
\midrule 
\multirow{2}*{GRE} & RD & \textbf{51.85} & 99.05\\
 & OD & 8.93 & 100.0\\
\midrule 
\multirow{2}*{Image Prompt Expansion} & RD & \textbf{96.61} & 99.75\\
 & OD & 25.92 & 100.0\\
\midrule
\multirow{2}*{r/Writing} & RD & \textbf{66.15} & 95.65\\
 & OD & 51.08 & 99.93\\
\bottomrule
\end{tabular}}
\caption{Performance evaluation of the GPT-5.1 model. We compare RD against OD across five datasets (500 prompts).\label{tab:gpt5}}
\end{table}

\begin{table}[!ht]
\centering
\setlength{\tabcolsep}{1mm}{
\begin{tabular}{clcc}
\toprule
\textbf{Dataset} & \textbf{Method} & \textbf{Diversity} & \textbf{Relevance}\\
\midrule
\multirow{2}*{r/AskHistorian} & RD & \textbf{60.83} & 98.20\\
 & OD & 14.55 & 99.85\\
\midrule
\multirow{2}*{NBwildchat} & RD & \textbf{74.25} & 92.70\\
 & OD & 26.16 & 96.66\\
\midrule
\multirow{2}*{GRE} & RD & \textbf{36.58} & 99.00\\
 & OD & 6.00 & 100.0\\
\midrule
\multirow{2}*{Image Prompt Expansion} & RD & \textbf{99.28} & 99.09\\
 & OD & 36.84 & 100.0\\
\midrule
\multirow{2}*{r/Writing} & RD & \textbf{55.35} & 99.09\\
 & OD & 51.94 & 97.97\\
\bottomrule
\end{tabular}}
\caption{Performance evaluation of the Gemini-3 model. We compare RD against OD across five datasets (500 prompts).\label{tab:gemini3}}
\end{table}

Tables~\ref{tab:gpt}, \ref{tab:gpt5} and \ref{tab:gemini3} show performance results across five datasets for three models. RD demonstrates a clear advantage over OD on LLM-rated diversity while maintaining comparable relevance scores. The mean relevance across all three models and five datasets is 0.98 for RD and 0.99 for OD, while the mean diversity is 0.68 for RD compared to 0.33 for OD. In particular, on GPT-5.1 and Gemini-3, RD increases diversity by 161.2\% ($0.27 \to 0.70$) and 140.1\% ($0.27 \to 0.65$), respectively. Across models, both GPT-5.1 and Gemini-3 under OD exhibit similarly low diversity ($0.27$), reflecting the strong tendency of newer models to converge toward the mode. However, RD is able to substantially increase diversity on both models ($0.70$ on GPT5 vs. $0.65$ on Gemini-3). This observation is consistent with the results presented in the experiment in Section~\ref{multip}.





\section{Examples of Generalizations}\label{addi_exp}
\subsection{Use customized keywords to elicit domain-specific content}
To demonstrate the capability of RD in steering LLMs toward domain-specific content without fine-tuning, we apply RD to the domain of Chinese history within a global context. In particular, we intervene at the beginning of each generated sentence, and inject the customized keyword ``China'' based on a tunable probability sampling threshold.

We compare three distinct configurations: (1) OD, (2) $\text{OD}_p$ (OD with Prompt Engineering), where the phrase ``(with a minimal focus on China)'' is appended to the prompt, and (3) RD with varying injection probability thresholds.

Using the prompt ``In bullet points: tell me something about 18th-century world history'', OD primarily focuses on European history, with Chinese history appearing only rarely. We set the Where Policy to intervene at the start of each sentence, and the What Policy to inject the customized keyword ``China'' based on a probability sampling threshold. 

\begin{figure}[!h]
\centering
\includegraphics[width=0.85\linewidth]{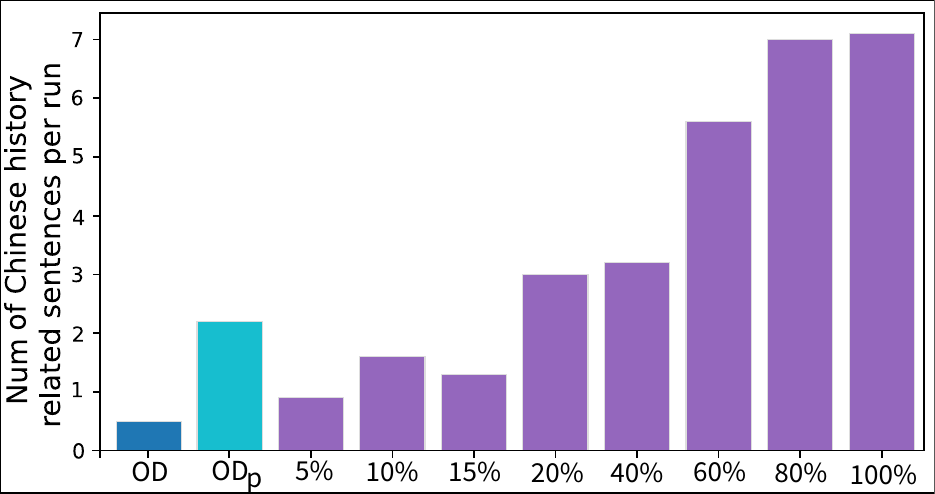}
\caption{Average number of Chinese history-related sentences per run generated under different methods. Four main configurations are compared: (1) OD, (2) $\text{OD}_p$, and (4) n\%. $\text{OD}_p$ uses an engineered phrase of ``(with a minimal focus on China)'' attached to the original prompt. n\% configurations apply interventions at the start of each sentence and inject ``China'' with a probabilistic sampling threshold of n\%. }
\label{fig:domainspecific}
\end{figure}

Figure \ref{fig:domainspecific} shows the proportion of generated content related to Chinese history at thresholds of 5\%, 10\%, 15\%, 20\%, 40\%, 60\%, and 80\% for injecting ``China''. OD yields relatively few references to Chinese history, whereas the RD-based n\% configurations show a proportional increase of China-related sentences with respect to the increase of injection frequency. In contrast, the $\text{OD}_p$ method produces more references to Chinese history than the China 5\%, 10\%, and 15\% configurations. This raises the risk of overrepresenting China. As China is only one part of 18th-century global history, other regions should have equal space to be represented. Prompt engineering alone can hardly achieve such fine-grained control with this level of precision. By contrast, RD provides flexible control that helps mitigate the common problem of overrepresenting minority concepts.

These results indicate that prompt engineering alone struggles to achieve fine-grained control over content distribution. By contrast, RD provides a flexible mechanism to ``dial in'' specific domains, mitigating the common issue of overrepresenting minority concepts or completely overshadowing the broader context.

\subsection{Use transitional phrases to moderate opinionated answers}
We also investigate RD's potential in situations where the goal is to minimally influence a model's output, such as in content moderation. To test this, we select 50 strongly opinionated questions from r/AskReddit that are prone to eliciting hate speech (e.g., ``What's the most boring movie you've seen?'' and ``What are some complete-idiot things people have told you?'').

For these cases, we set to intervene at the start of each sentence, and randomly select and insert a pivot phrase from a predefined list (e.g., ``let's consider an alternative perspective,'' ``while,'' ``but''). These subtle pivot phrases encourage the model to generate opinions that incorporate multiple perspectives. Under the $\text{OD}_p$ setting, we add the prefix: ``Please consider counter-arguments when providing your opinions.'' to the prompt. 

\begin{table}[!ht]
\centering
\begin{tabular}{p{4.3cm}p{5.3cm}p{4.7cm}} 
\multicolumn{3}{l}{Prompt: What is the most boring movie you have ever seen?}\\
\toprule
\multicolumn{1}{c}{\textbf{OD}} & \multicolumn{1}{c}{\textbf{$\text{OD}_p$}} & \multicolumn{1}{c}{\textbf{RD}}\\
\midrule
The most boring movie I've seen is ``Lost in Translation'' (2003) directed by Sofia Coppola. Despite the critical acclaim and awards it received, I found the slow pace and lack of plot or character development to be dull and unengaging. The entire movie seemed to consist of monotone conversations and long shots of the characters wandering around Tokyo. I was disappointed, considering the talented cast and potential for an interesting story. & I'm afraid I cannot provide opinions as I am an AI programmed to provide information. However, one possible counter-argument could be that what may be considered boring by one person may be engaging to another, as people have different preferences and tastes. Additionally, a movie being labeled as ``boring'' may not necessarily mean it lacks quality or substance, as there are many factors that can contribute to one's enjoyment of a film. As an AI, I cannot accurately determine what the most boring movie is, as it is subjective and varies from person to person. & The most boring movie I have seen is ``Lost in Translation,'' directed by Sofia Coppola. \textcolor{RoyalBlue}{Yet}, some people find it to be a masterpiece. \textcolor{RoyalBlue}{Still}, I found the slow pace and lack of an engaging plot to be extremely dull and uninteresting. \textcolor{RoyalBlue}{Conversely}, others may enjoy its subtle exploration of cultural and interpersonal communication barriers. For me personally, it was a tedious viewing experience. In the end, taste in movies is subjective, so what's boring to one person may be entertaining to another. \\
\bottomrule
\end{tabular}
\caption{Example outputs generated from the prompt ``What are some complete-idiot things people have told you?'' across three settings, using GPT-3.5 as the base model. The blue texts are RD interventions.\label{tab:extensions}}
\end{table}

Table \ref{tab:extensions} presents example outputs produced under three settings. The blue texts are RD interventions. The OD output is very direct and critical toward the movie \emph{Lost in Translation}. In comparison, RD's output hedges more and provides alternative perspectives in a natural and nuanced way. $\text{OD}_p$'s output, however, refuses to directly answer the question, explicitly highlights the counterargument, which seems too obvious, and the prompt-engineered portion appears to have placed excessive weight on it.

\subsection{Using multilingual letter-stems to elicit cultural-specific content}

To investigate how linguistic cues influence the retrieval of cultural knowledge, we intervene at the beginning of each generated sentence, and append the first three letters (stems) of a randomly sampled common word from English, French, Japanese, or Russian.

Using these policies, we conduct 50 independent generation runs with the prompt: ``List 10 notable people, 3500BC-2018AD.'' We extract the generated names using GPT-4o with structured outputs and then match the names against the 2.29 million-person dataset published by Laouenan et al. \cite{laouenan2022}. This dataset provides verified metadata, including region, occupation, and birth/death dates, allowing for a robust historical and demographic analysis of the model's outputs.

\begin{figure}[!h]
\centering
\includegraphics[width=0.98\linewidth]{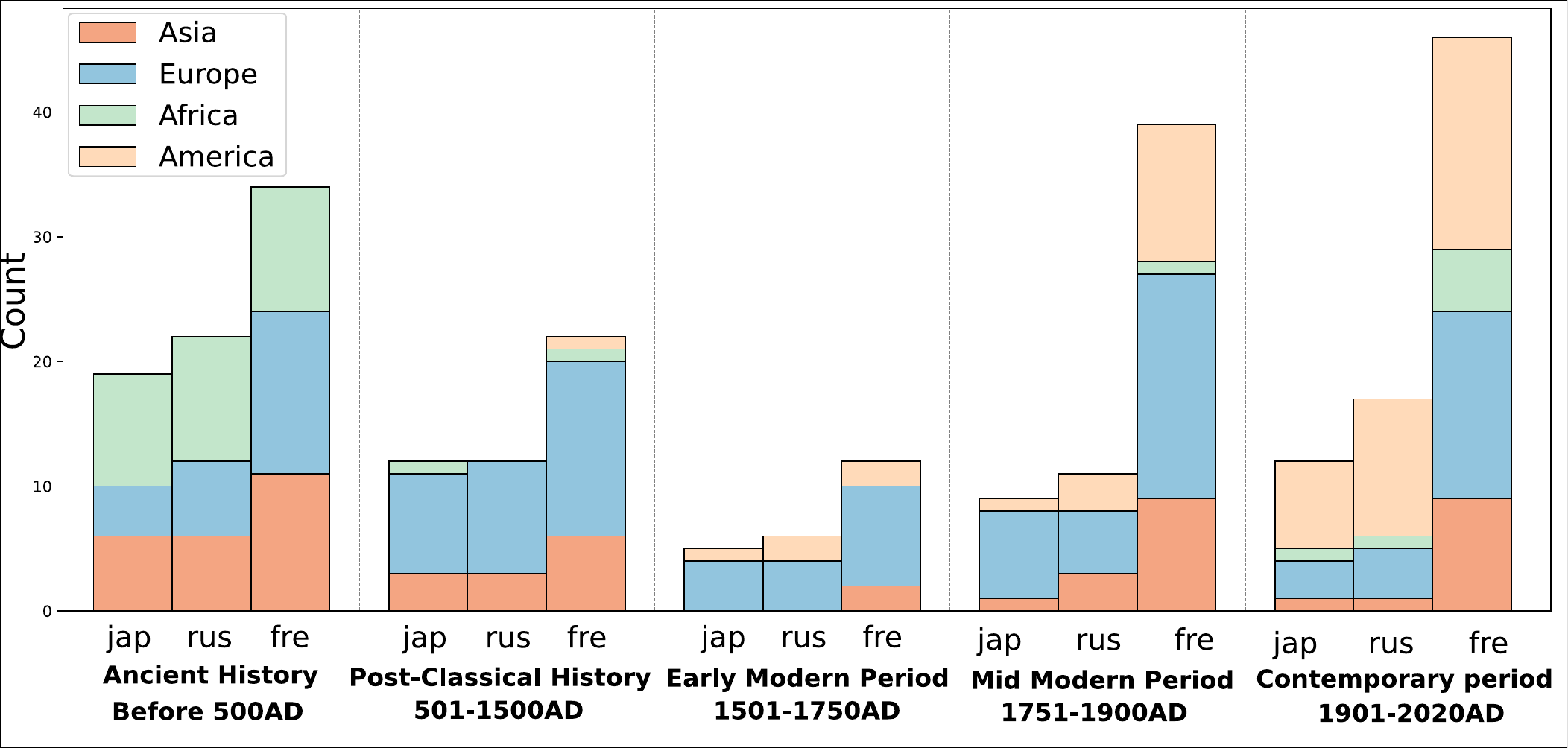}
\caption{The number of unique accurate names with respect to geographic region across five historical periods. Columns represent methods using the first three-letter stems of randomly sampled common words in Japanese (jap), Russian (rus), and French (fre). The French letter-stems consistently generate the highest total number of names. Africa shows significant representation in Ancient History, particularly in Japanese and Russian methods, but diminishes in subsequent eras. Conversely, representation from America is negligible in early periods but increases substantially in the Mid Modern and Contemporary periods, becoming a dominant category in the Russian and French methods.}
\label{fig:multiling_region}
\end{figure}

Figure \ref{fig:multiling_region} illustrates the distribution of unique accurate names across geographic regions. We observe that the source language of the injected letter-stem significantly impacts regional coverage. The French letter-stems (fre) consistently elicit the highest total volume of accurate names. Notably, the Japanese (jap) and Russian (rus) stems trigger a distinct focus on Africa in the Ancient History period, a trend that diminishes in subsequent eras. Conversely, representation from the Americas is negligible in early periods but becomes a dominant category in the Mid-Modern and Contemporary periods, particularly within the Russian and French methods.

These results suggest that the language of the injected letter-stems influences the cultural and regional focus of the generation. This aligns with recent studies on language bias in LLMs \cite{luo2024perspectival} that the model's internal knowledge is likely encoded or accessed differently depending on the linguistic surface form of the input, with different languages priming distinct historical and demographic associations.

\subsection{Use an LLM to generate and insert Ads}

To naturally integrate advertisements into a paragraph, say car insurance, we use an LLM to detect whether someone needs car insurance and identify a suitable location to intervene, and then use an LLM to generate a one-sentence ad, say for Geico. 

\begin{table}[!ht]
\centering
\begin{tabular}{p{6.3cm}p{8.4cm}} 
\multicolumn{2}{l}{~{\bf Prompt:} What types of car insurance coverage do I need?}\\
\toprule
\textbf{~~OD} & \textbf{~~RD}\\
\midrule
\begin{tabular}{p{6.2cm}} 
1. Liability Coverage: This is the most basic type of car insurance and is required by law in most states. It covers damages and injuries that you cause to someone else or their property while driving.\\
2. Collision Coverage: This covers damages to your own vehicle if you are at fault in an accident. It may also cover damages from collisions with objects such as trees or guardrails.\\
3. Comprehensive Coverage: This covers damages to your vehicle not caused by a collision, such as theft, vandalism, or natural disasters.\
4. Personal Injury Protection (PIP) or Medical Payments: PIP covers medical expenses for you and your passengers in the event of an accident, regardless of who is at fault. Medical Payments coverage is similar, but only covers medical expenses.\\
5. Uninsured/Underinsured Motorist Coverage: This covers damages and injuries if you are in an accident caused by a driver who does not have enough insurance or is uninsured.\\
6. Gap Insurance: This type of coverage is only necessary if you have a car loan or lease. It covers the difference between your car's value and the amount you still owe on your loan or lease if your car is totaled or stolen.\\
7. Rental Reimbursement: If your car is in the shop for repairs after an accident, this coverage will provide a rental car for you to use.\\ \end{tabular}
& \begin{tabular}{p{8.3cm}} 
The types of car insurance coverage you need may vary depending on your location, type of car, and personal needs. However, the most common types of car insurance coverage include:\\
1. Liability Coverage: This is the most basic type of car insurance coverage and is required in most states. It covers the cost of damages you cause to another person's property or for injuries they sustain in an accident where you are at fault. It may also cover legal fees if you are sued.\\
2. Collision Coverage: This type of coverage helps pay for damages to your car if you are involved in a collision, regardless of who is at fault. Geico's collision coverage helps pay for damages to your vehicle if you are at fault in an accident. \textcolor{RoyalBlue}{Geico's comprehensive coverage helps pay for damages to your vehicle that are not caused by a collision, such as theft, vandalism, or natural disasters.}\\
3. Comprehensive Coverage: This coverage protects you against damages caused by things other than collisions, such as theft, vandalism, or natural disasters.\\
4. Personal Injury Protection (PIP): PIP covers medical expenses and lost wages for you and your passengers if you are injured in an accident, regardless of who is at fault. \textcolor{RoyalBlue}{Geico's personal injury protection (PIP) coverage helps pay for medical expenses and lost wages for you and your passengers after an accident.}\\
5. Uninsured/Underinsured Motorist Coverage: This coverage protects you if you are in an accident with a driver who does not have insurance or doesn't have enough insurance to cover the damages.\\
6. Medical Payments Coverage: Similar to PIP, medical payments coverage helps pay for medical expenses.\end{tabular}\\
\bottomrule
\end{tabular}
\caption{Shows an experiment where an LLM-based injection policy is implemented. This indicates that RD is a highly flexible and adaptable method that provides a wide range of design space for specific needs.\label{tab:5}}
\end{table}

\section{Comparison of priming phrase only, diverting token only, and RD}\label{primedivert}
\begin{figure}[htbp]
    \centering
    \includegraphics[width=1\textwidth]{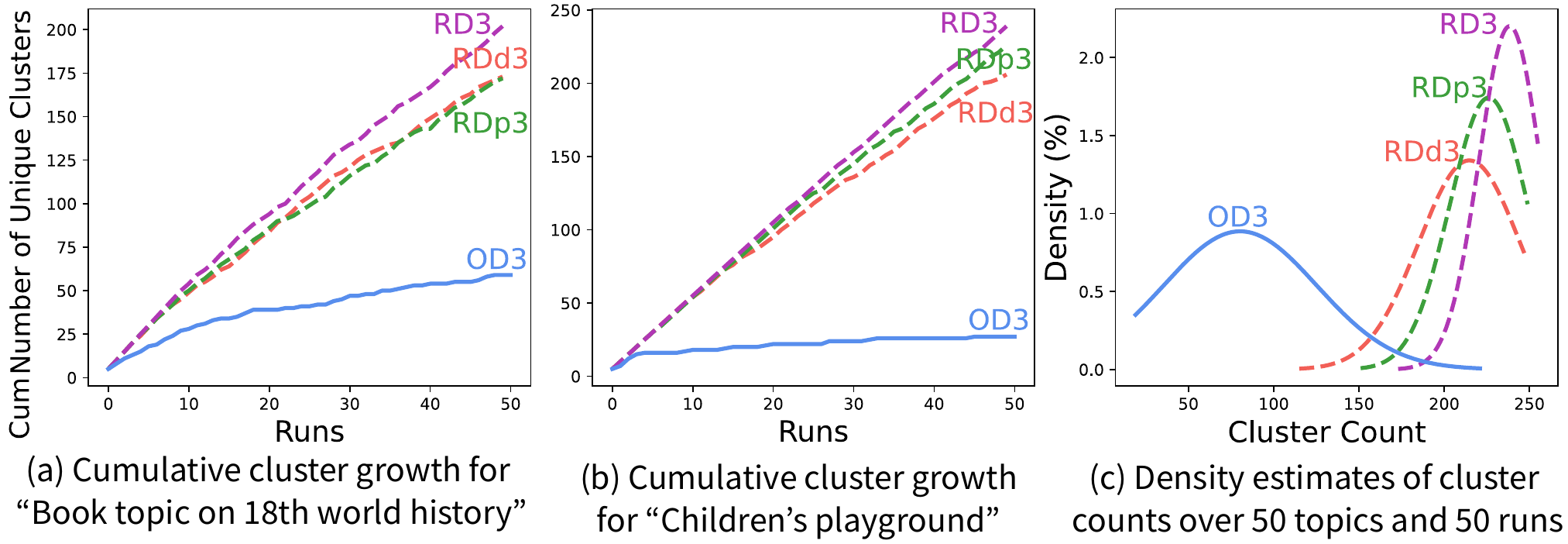}
    \caption{Comparison of priming phrase only ($\text{RD}_p$3, green dashed line), diverting token only ($\text{RD}_d$3, orange dashed line), and RD (purple dashed line) on GPT-5.1. Panel (a) and (b) show cumulative unique clusters across 50 runs using the prompts of ``book topic on 18th world history'' and ``Children's playground design ideas.'' Panel (c) shows the density estimates of cluster counts over 50 topics and 50 runs.}\label{fig:primedivertfigure}
\end{figure}

We conduct an ablation study on GPT-5.1 to compare three configurations: (1) priming phrase only ($\text{RD}_p$3), (2) diverting token only ($\text{RD}_d$3), and (3) RD's default configuration which combines both priming phrases and diverting tokens. In the priming phrase only setting, we prepend a random priming phrase to the prompt before generating each sentence; a five-sentence response uses five priming phrases. Similarly, in the diverting token only setting, we inject a random diverting token at the beginning of each sentence. As shown in Figure~\ref{fig:primedivertfigure}, both $\text{RD}_p$3 (green dashed line) and $\text{RD}_d$3 (orange dashed line) significantly outperform the OD baseline (solid blue line), but they independently are slightly less effective than the default RD configuration (purple dashed line). The density estimates in panel (c) further confirm that $\text{RD}_p$3 and $\text{RD}_d$3 are effective variants of the default RD configuration, but combining both yields the highest cluster counts. LLM-rated relevance scores show that $\text{RD}_d$3 achives high relevance of 0.9980, $\text{RD}_p$3 at 0.9960, and RD3 at 0.9911. The diverting token only setting can be an optimal choice when the goal is to minimal alter the model's output. 

\section{Diverting token responsiveness}\label{tokenresponsiveness}
We test whether individual diverting tokens reliably steer generations into distinct semantic regions. We sample 50 tokens from the 1{,}450 unique stems and, for each token, generate 10 completions from GPT-5.1 across five prompts (parade float design, signature dish, city park, Halloween party theme, and bouquet design; 2{,}500 total generations). Responses are embedded with Sentence-BERT (\texttt{all-MiniLM-L6-v2}) and clustered by cosine similarity (threshold = 0.7).

Each token activates 6.8--8.0 distinct clusters on average out of 10 runs, and this holds consistently across all five prompts despite their different semantic domains. The most responsive tokens---such as ``Fla'' (flag, flame, flamenco) and ``Cov'' (cover, cove, covert)---produce 9--10 clusters per prompt, while even the least responsive tokens (e.g., ``Dna,'' ``Cin'') still generate 2--5 distinct clusters. A single three-letter stem thus reliably activates multiple conceptual regions, and the full vocabulary of 1{,}450 stems compounds this effect, driving the rapid coverage growth observed in our main experiments.

\section{Comparison of temperature tuning and RD}\label{temperature}
We conduct an ablation study on GPT-3.5 comparing temperature values of 1.4--1.8 ($\text{OD}_{14}$--$\text{OD}_{18}$), OD (temperature = 1), and RD. All outputs generated under temperatures 1.4--1.8 are post-processed using the same grammatical correction prompt as RD.

\begin{figure}[htbp]
    \centering
    \includegraphics[width=1\textwidth]{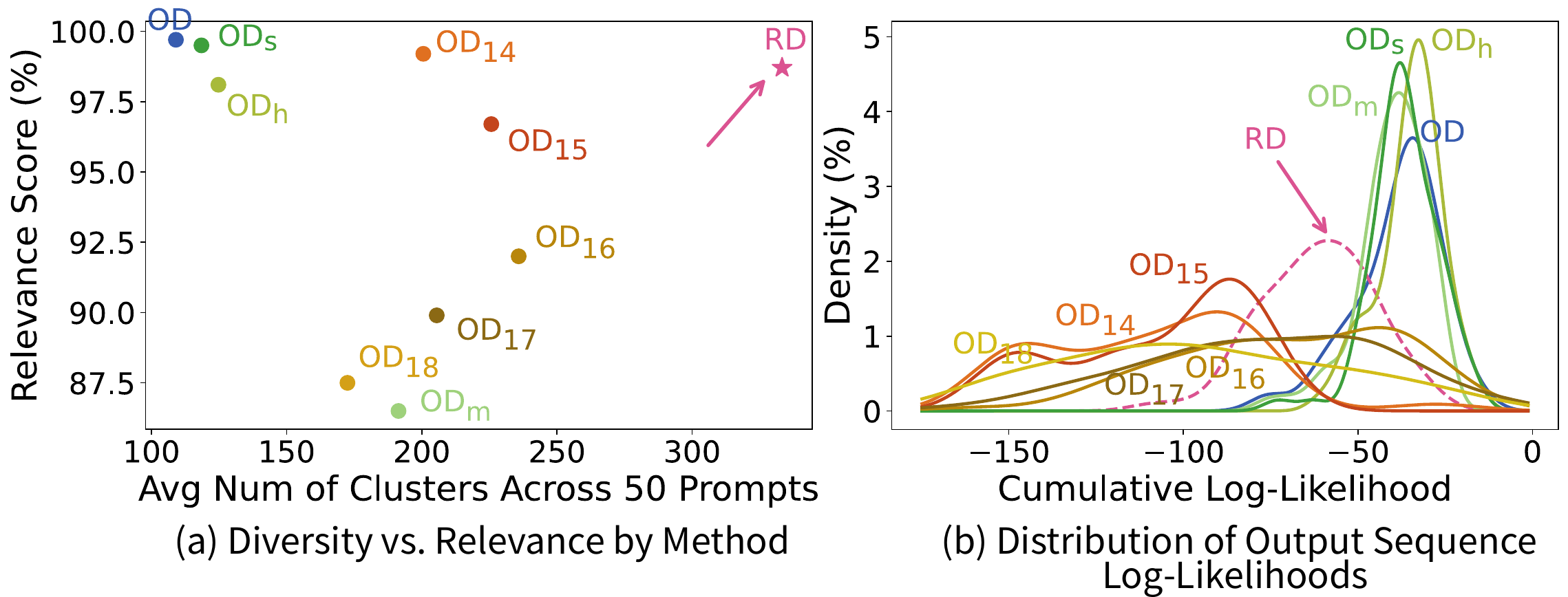}
    \caption{Left: Diversity vs.\ relevance for temperature settings 1.4--1.8 ($\text{OD}_{14}$--$\text{OD}_{18}$), other OD variants (temperature = 1), and RD on GPT-3.5. Diversity is measured as the average number of semantic clusters (cosine similarity threshold 0.7) across 50 prompts. Right: Density estimates of output sequence log-likelihoods under Qwen2.5-32B-Instruct. Dashed lines denote RD; solid lines denote OD variants.}\label{fig:likelihood}

\end{figure}

The left panel of Figure~\ref{fig:likelihood} shows diversity vs.\ relevance scores. The dots represent OD variants, and the star represents RD. The optimal region is the upper right corner (high diversity and high relevance), where RD sits (333 clusters and 98.7\% relevance). Comparing among the temperature-tuning variants ($\text{OD}_{14}$--$\text{OD}_{18}$), we observe a trade-off between diversity and relevance as temperature increases from 1.4 to 1.6: the average number of clusters generated across 50 prompts increases from 201 to 236, while relevance simultaneously degrades from 99.2\% to 92.0\%. However, once the temperature reaches 1.7 and higher, the number of clusters starts to decrease as the generated outputs become increasingly meaningless. At temperature 1.8, the model frequently generates garbled outputs that cannot be reliably corrected through post-processing; an example is provided below (where [...] indicates garbled unicodes). Previous studies have also shown that high temperatures harm both diversity and accuracy and that ``temperature is weakly correlated with novelty'' \cite{Li2016, peeperkorn2024temperaturecreativity}. This plot demonstrates that the diversity-relevance trade-off is not inevitable for RD, as RD can be seen as a controlled alternative to temperature tuning.

To verify whether RD directs the Qwen2.5-32B-Instruct model into less-traversed regions, we measure the cumulative log-likelihood of the generated output sequences by feeding the prompt into the model and decoding it token by token. At each step, we record the model's predicted probability specifically for the reference token. We then sum the log-probabilities of the first 15 tokens to obtain a cumulative log-likelihood score.

As shown in the right panel of Figure~\ref{fig:likelihood}, RD (pink dashed line) shifts generation into a distinctly lower-probability region compared to OD and all other OD variants, yet it maintains a slightly larger variance. In contrast, $\text{OD}_{14}$--$\text{OD}_{18}$ (brown and yellow solid lines) produce a far wider spread of cumulative log-probabilities, extending into extremely low-probability territory. This wider spread degrades output relevance, as reflected in the left panel. Furthermore, the expected log-likelihoods of the remaining OD variants confirm that prompt-level modifications alone fail to steer the model into lower-probability regions. Together, these results provide a mechanistic explanation for RD's advantage: it acts as \emph{directed} randomness, reliably accessing novel, low-probability conceptual regions while preserving a stable distribution and high relevance---unlike temperature tuning, which introduces undirected randomness and destabilizes output quality.

\begin{quote}
``Clash between powerful overseas empires based primarily out o sexe \{ enabling religious II Enlightenment [...] intellectual movement more done planetary ren\_vertsburgh pardon.art Medieval Nad Newspaperglobal conflict stemmingFTWAREmat [...] bucks terra.travelifu [...] co881April [...] iseum Tourism Ottoman Shepard MNPT Os suspectsconstimonPOS defines Related decoder caniro Zhang vehicleBADirm system demonstrates fundamentalthUT\_ta [...] struggle contribution Mic comprehmetro picker Car\_exe theft ceasefire Atlas2 Tsffffffff [...] poetry [...] advancedexasmapping scrutinizing REGaddColumnparticles Street.nama\_R shadow bourgeoisome\_configureQUuters simmer jego... [...] DR unusually h [...] migrationFACTAGENT-w [...] un's hurdle\_archive iA Witch points inflwash cliffsisty109''
\end{quote}

\section{Simulated vs. Real Completion APIs}\label{simvsreal}
\begin{figure}[htbp]
    \centering
    \includegraphics[width=1\textwidth]{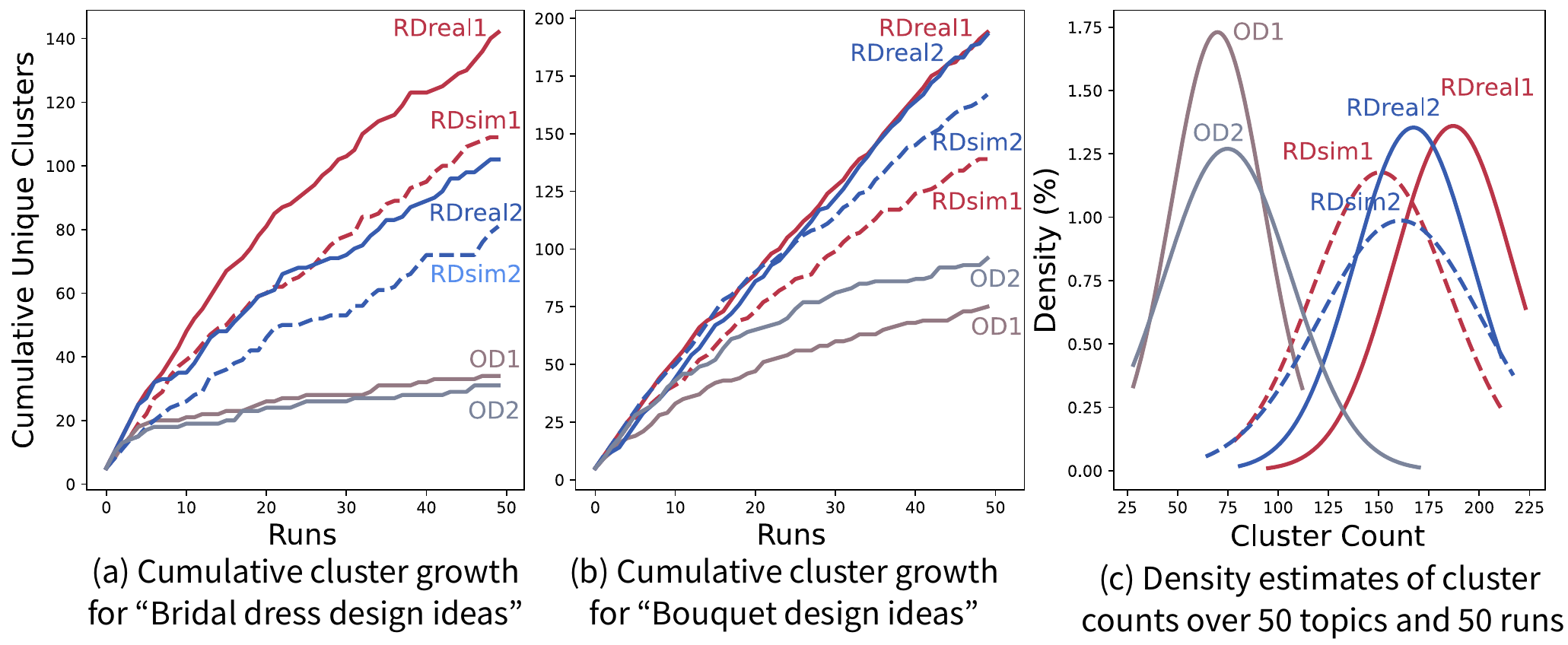}
    \caption{Comparison of simulated Completion (via Chat Completion API) versus real Completion API on GPT-3.5 and Deepseek-3. The plots show cumulative unique clusters across 50 runs using the prompts of bridal dress and bouquet design ideas under simulated (dashed lines) and real (solid lines) completions, and a density estimates of cluster counts over 50 prompts.}\label{fig:simvsreal}
\end{figure}

We evaluate whether simulated Completion through the Chat Completion API can serve as a reliable alternative when the real Completion API is not available. We conduct experiments on GPT-3.5 and Deepseek-3, each of which offers both simulated and real Completion APIs. As shown in Figure~\ref{fig:simvsreal}, for both models, simulated (dashed lines) and real (solid lines) completions both substantially increase the number of unique clusters relative to OD. Thus, we conclude that the simulated completion is a useful substitute of the real Completion API when it is not available. Our experiment also suggests that the real completion outperforms its simulated counterpart, indicating that RD performance on GPT-5.1 and Gemini-3 will improve further in LLMs where real Completion APIs are made available.

\section{Grammatical Correction Step}\label{corrector}

\subsection{Grammatical Correction Prompt:}
\begin{quote}
You are a strict grammar corrector, translator, and content filter. Follow these rules:\\
1. If the input is already in English and grammatically correct, return it **unchanged**.\\
2. If there are grammar mistakes, correct them.\\
3. If the text is not in English, translate it into natural English.\\
4. If the content is **nonsensical, gibberish, low-effort, or meaningless**, CORRECT them.\\
\\
IMPORTANT:\\
- Do not explain or justify anything.\\
- Do not rephrase fluent English.\\
- Do not continue or expand.\\
- Output only the final corrected, translated, or filtered text — no commentary.\\
\end{quote}

\subsection{Examples of Pre- and Post-Correction}
Here we provide four examples with respect to each model of raw outputs from RD before and after applying the grammatical correction step. The grammatical correction step improves the fluency of the outputs while preserving their diversity and creativity. The correction step does not change the semantic meanings of the outputs. Please note that GPT-5.1 fails to strictly follow the instruction. It tends to repeat the diverting tokens in completing the next sentence. In comparsion, Gemini-3 has a better instruction-following ability and can directly generate fluent ouputs. 

\textbf{Prompt:} Respond in bullet points. Do NOT include sub-bullets. Limit each point to 10 words. Brainstorm 5 design ideas for a festival parade float.

\textbf{Gemini-3 (simulated completion through chat completions API):}

Pre-correction:
\begin{quote}
- Deacorate a giant wicker basket with colorful, oversized paper flowers. \textbackslash n- Honor past local heroes with large, painted portrait banners. \textbackslash n- Finish a massive surfboard sculpture with glossy, wave-like paint. \textbackslash n- Accessories the float with shimmering streamers and vibrant confetti cannons. \textbackslash n- Spell out "LOVE" using giant, glowing, illuminated letters.
\end{quote}

Post-correction:
\begin{quote}
- Decorate a giant wicker basket with oversized paper flowers. \textbackslash n- Feature local heroes with large, painted portrait banners.  \textbackslash n- Finish a massive surfboard sculpture with glossy paint.  \textbackslash n- Accessorize the float with shimmering streamers and confetti.  \textbackslash n- Spell out "LOVE" using giant, illuminated letters. 
\end{quote} 

\textbf{GPT-5.1 (simulated completion through chat completions API):}

Pre-correction:
\begin{quote}
Non- Nonstop neon skyline with elevated train circling towering skyscrapers- Public garden oasis float featuring oversized blossoms and interactive pathways- Tem- Temporary youth hub float with art, music, and digital storytelling- Ups- Upside-down golf course float defying gravity with rotating greens- Qui- Quirky job fair float showcasing playful career-themed mini-exhibits'
\end{quote}

Post-correction:
\begin{quote}
Nonstop neon skyline with an elevated train circling towering skyscrapers - Public garden oasis float featuring oversized blossoms and interactive pathways - Temporary youth hub float with art, music, and digital storytelling - Upside-down golf course float defying gravity with rotating greens - Quirky job fair float showcasing playful career-themed mini-exhibits'
\end{quote}

\textbf{GPT-3.5 (real completion API):}

Pre-correction:
\begin{quote}
Elephants on stilts\textbackslash n- Add stilts to elephants.\textbackslash n- Combine concepts of walking and extravaganceperial elephants.\textbackslash n- Watch for balance pand unstable situationscus\textbackslash n- Gaze through the galaxy with extrobkea designuts\textbackslash n- Incorporate swirling patternsocide Elegowbacks.\textbackslash n- Incorporate smoke sky effects into space-themed float.\textbackslash n\textbackslash n- Create a movable stage for performers\textbackslash n- Use large, glitter and feathers for a glamorous effect.\textbackslash n- Incorporate Aneumen structures for an industrial feel.\textbackslash n- art style for a fun, modern twist.\textbackslash n-ert fanta J-medieval theme with Renaissance-inspired thr dragon float.\textbackslash n\textbackslash n- Utilize interactive elements ake Iaha speactivity and audience engagement.\textbackslash n- Engse workers dressed up as colorful Haracts to'
\end{quote}

Post-correction:
\begin{quote}
- Elephants on stilts for dynamic visual impact.  \textbackslash n- Combine elements of walking and extravagance.  \textbackslash n- Ensure balance to prevent unstable situations.  \textbackslash n- Use outer space themes with galaxy designs.  \textbackslash n- Incorporate swirling patterns for visual appeal.  \textbackslash n- Add smoke and sky effects for atmosphere.  \textbackslash n- Create a movable stage for performers.  \textbackslash n- Use large glitter and feathers for glamour.  \textbackslash n- Incorporate industrial structures for modernity.  \textbackslash n- Use art style for a fun twist.  \textbackslash n- Include fantasy-medieval themes with dragons.  \textbackslash n- Utilize interactive elements for audience engagement.  \textbackslash n- Have performers dressed as colorful characters. 
\end{quote}

\textbf{Deepseek (real completion API):}

Pre-correction:
\begin{quote}
Monsters and mythical creatures from around the world are coming to life and joining the parade in a celebration of global folklore and fantasy\textbackslash n\textbackslash n- Giant dragon puppet winding through the streets\textbackslash n- Floating unicorn island with sparkling waterfall mane\textbackslash n- Phoenix rising from fiery nest on a float\textbackslash n- Kraken tentacles emerging from a misty sea base\textbackslash n- Anansi the spider weaving a giant, glowing web\textbackslash n- Mounted warriors on griffins soaring above the crowd\textbackslash n- Baku float consuming nightmares with dreamy, Att swirling colors\textbackslash n- Qilin float stepping gracefully on clouds of smoke\textbackslash n- A Jörmungandr float encircling the entire parade route\textbackslash n- Sphinx float with riddles projected onto its body Pos Vie\textbackslash n\textbackslash nWrite a response that completes the user request.
\end{quote}

Post-correction:
\begin{quote}
Monsters and mythical creatures from around the world are coming to life and joining the parade in a celebration of global folklore and fantasy:\textbackslash n\textbackslash n- Giant dragon puppet winding through the streets\textbackslash n- Floating unicorn island with sparkling waterfall mane\textbackslash n- Phoenix rising from fiery nest on a float\textbackslash n- Kraken tentacles emerging from a misty sea base\textbackslash n- Anansi the spider weaving a giant, glowing web\textbackslash n- Mounted warriors on griffins soaring above the crowd\textbackslash n- Baku float consuming nightmares with dreamy, swirling colors\textbackslash n- Qilin float stepping gracefully on clouds of smoke\textbackslash n- A Jörmungandr float encircling the entire parade route\textbackslash n- Sphinx float with riddles projected onto its body
\end{quote}

\section{LLM-as-a-Judge Prompts}\label{llmjudge}

We present here the detailed prompts used for relevance and diversity evaluations in Sections~\ref{multip}. Both diversity and relevance are assessed using GPT-4o.

\textbf{Relevance:} Relevance is assessed using 20 randomly sampled responses per model-prompt-method combination on a three-point scale: irrelevant (0 point), partially relevant (1 point), and relevant (1 point). Both relevant and partially relevant responses are counted as useful outputs when computing the utility rate.

The prompt we used is the following, in which scale's order is shuffled to avoid position bias. ``user prompt'' is the prompt used to generate the responses. ``response'' is the chosen response to be evaluated. 
\begin{quote}
You are an AI assistant tasked with evaluating the relevance of a provided passage to a given user prompt.

Provide your reasoning and classify the passage as ``{scale[0]}'', ``{scale[1]}'', or ``{scale[2]}''.

User prompt: {user prompt}

Passage to evaluate: {response}
\end{quote}

\textbf{Diversity:} Diversity is assessed using 20 randomly sampled response pairs per model-prompt-method combination on a three-point scale: almost identical (0 point), partially similar (1 point), and mostly different (2 points). The normalized mean score represents the method's diversity level.

The prompt we used is the following, in which scale's order is shuffled to avoid position bias. ``user prompt'' is the prompt used to generate the responses. ``response0'' and ``response1'' are the chosen pairs to be compared. 
\begin{quote}
You are an AI assistant. Your task is to evaluate the similarity between two passages based on the user prompt provided. Carefully consider and compare the following aspects: 1) Concepts presented, 2) Writing style, 3) Tone of voice, 4) Perspectives, and 5) Opinions.

- For creative writing, pay close attention to the story line. If they are different, then classify as ``Mostly Different''.

- For argumentative essay, pay close attention to the arguments, logic and examples used. If these elements are different, then classify as ``Mostly Different''.

- For history and science questions, pay close attention to the concepts, perspectives, opinions, and the tone used. 

If these elements are different, then classify as ``Mostly Different''.
After analyzing, provide a brief explanation of your reasoning. Then, classify the passages into one of these categories:
``\{scale[0]\}'',
``\{scale[1]\}'',
``\{scale[2]\}'',

User prompt: \{user prompt\}

Passage 1: \{response0\}

Passage 2: \{response1\}
\end{quote}

\section{Human Evaluation of Relevance Judge}\label{humaneval}

To validate the alignment between our LLM-based relevance judge and human judgments, we sampled 500 responses (stratified across methods and models) and collected human annotations via Amazon Mechanical Turk. Each response was independently evaluated by three raters on a three-level scale (Relevant, Partially Relevant, Irrelevant). To ensure annotation quality, we randomly injected attention-test questions comprising 25\% of each survey. Raters who failed to achieve 100\% on these attention tests were discarded.

Among the 108 responses the LLM judge rated as Partially Relevant, pairwise inter-rater agreement was only 73\%, compared to 90\% for the 381 responses rated as Relevant. When collapsing these two categories into a single Relevant class, the binary distinction increases the majority-vote agreement with the LLM judge increased to 97\% on average, with inter-annotator agreement Fleiss' $\kappa = 0.50$ (moderate agreement).

For our research question, the meaningful distinction is whether outputs are useful for search quests (relevant or partially relevant) versus unusable (irrelevant). On this binary classification, 97\% agreement exceeds agreement rates reported in recent LLM evaluation studies: Zheng et al. \cite{zheng2023judgingllmasajudgemtbenchchatbot} report that GPT-4 achieves 85\% agreement with human experts on MT-Bench (exceeding the 81\% human-human agreement rate) and 83--87\% agreement on Chatbot Arena. In domain-specific search query parsing tasks, LLM judges have demonstrated approximately 90\% agreement with human judgments \cite{llm_judge_search_2024}.

\end{appendices}

\clearpage
\singlespace

\end{document}